\documentclass[journal]{IEEEtran}
\newcommand{\papertitle}{
Multi-Mapping Image-to-Image Translation with Central Biasing Normalization
}
\usepackage[abspath]{}

\usepackage{capt-of} 
\newcommand{\stimes}{{\times}}

\newcommand*{\defeq}{\mathrel{\vcenter{\baselineskip0.5ex \lineskiplimit0pt
                     \hbox{\scriptsize.}\hbox{\scriptsize.}}}%
                     =}

\usepackage[noadjust]{cite}
\usepackage{graphicx} 
\graphicspath{ {./pdf/} }
\DeclareGraphicsExtensions{.pdf} 

\usepackage{lipsum}

\usepackage{threeparttable}
\usepackage{bm}  % 斜体加粗体
\usepackage{esvect} % vv 向量箭头 https://tex.stackexchange.com/questions/269644/vector-bold-and-arrow-sign-on-top
\usepackage{subfigure}
\usepackage{soul} % st 刪除线
\makeatletter
\def\input@path{{../}{./}}
\makeatother

\usepackage{amsmath,amssymb} % define this before the line numbering.
\usepackage{color}
\usepackage[misc]{ifsym}
\usepackage[utf8]{inputenc}
\usepackage[T1]{fontenc}

\usepackage{xspace}

\makeatletter
\DeclareRobustCommand\onedot{\futurelet\@let@token\@onedot}
\def\@onedot{\ifx\@let@token.\else.\null\fi\xspace}

\def\eg{\emph{e.g}\onedot} 
\def\ie{\emph{i.e}\onedot}

\def\etal{\emph{et al}\onedot}
\makeatother

\ifdefined\IEEEkeywords

	\makeatletter
	\def\endthebibliography{%
	  \def\@noitemerr{\@latex@warning{Empty `thebibliography' environment}}%
	  \endlist
	}
	\makeatother
\else 
\fi

\ifdefined\IEEEPARstart
\else
	\newcommand{\IEEEPARstart}[2]{#1#2}
\fi
\ifdefined\appendices
\else
	\newcommand{\appendices}{\section*{Appendices}}
\fi

\newcommand{\latentcode}{c}

\newcommand{\bias}{b}
\newcommand{\affine}{f}

\newcommand{\Emean}[1]{\mu({#1})}
\newcommand{\variance}[1]{\sigma(#1)}

\usepackage{ stmaryrd } % \rrbracket \llparenthesis 各种框
\usepackage{booktabs}
\usepackage{multirow}

\newcommand{\inlinecode}{\texttt}

\newcommand{\ifjournal}[2]{
\ifdefined\IEEEkeywords #1
\else #2
\fi}

\usepackage{autobreak} 
\usepackage[colorlinks,linkcolor=blue]{hyperref} %超链接

\begin{document}
\title{\papertitle} 
\author{
        Xiaoming~Yu, 
        Zhenqiang~Ying,
        Thomas~Li,
        Shan~Liu,
        and~Ge~Li~\textsuperscript{\Letter},~\IEEEmembership{Member,~IEEE,}
\thanks{
This work was supported by the Project of National Engineering Laboratory for Video Technology-Shenzhen Division, Shenzhen Municipal Science and Technology Program under Grant (JCYJ20170818141146428), and National Natural Science Foundation of China and Guangdong Province Scientific Research on Big Data (No. U1611461).
This paper was recommended by Associate Editor X. XX.
}
\thanks{X.Yu, Z. Ying, T. Li, and G. Li are with the School of 
Electronic and Computer Engineering, Shenzhen Graduate School, Peking University, 
518055 Shenzhen, China (e-mail:
xiaomingyu@pku.edu.cn; zqying@pku.edu.cn; thomasli@pkusz.edu.cn geli@ece.pku.edu.cn).
S.Liu is with the Media Lab, Tencent (e-mail: shanl@tencent.com).
}
\thanks{Color versions of one or more of the figures in this paper are available
online at http://ieeexplore.ieee.org.
}
\thanks{Digital Object Identifier XX.XXXX/XXX.20XX.XXXXXXX
}
\thanks{Manuscript received XXX XX, 20XX; revised XXX XX, 20XX.}
}

\markboth{Journal of \LaTeX\ Class Files,~Vol.~14, No.~8, August~2018}%
{XXXXX \MakeLowercase{\textit{et al.}}: \papertitle}

\maketitle

\IEEEpeerreviewmaketitle

%!TEX root = bare_jrnl.tex

\begin{abstract}
Recent advances in image-to-image translation have seen a rise in approaches generating diverse images through a single network.
To indicate the target domain for a one-to-many mapping,
the latent code is injected into the generator network.
However, we found that the injection method leads to mode collapse 
because of normalization strategies.
Existing normalization strategies might either
cause the inconsistency of feature distribution or eliminate the effect of the latent code.
To solve these problems,
we propose the consistency within diversity criteria for designing the multi-mapping model.
Based on the criteria,
we propose central biasing normalization
to inject the latent code information.
Experiments show that our method can improve the quality and diversity of existing image-to-image translation models,
such as StarGAN, BicycleGAN, and pix2pix.
\end{abstract}


\section{Introduction}

\IEEEPARstart{M}{any} image processing and computer vision problems can be framed as image-to-image translation tasks~\cite{isola2017pix2pix},
such as facial synthesis~\cite{tian2018cr,taigman2017unsupervised,huang2018cartoon}, photo to sketch~\cite{zhang2015face,fanscoot}, and image colorization~\cite{zhang2016colorful}.
This can also be viewed as mapping an image from one specific domain to another.
Many studies have shown remarkable success in image-to-image translation between two domains,
\eg image synthesis~\cite{isola2017pix2pix}, inpainting~\cite{pathak2016context}, colorization~\cite{zhang2016colorful} and super-resolution~\cite{ledig2016photoresolution}. 
In these methods, the generative model tries to learn a specific mapping from the source domain to the target domain.
However,
these one-to-one mapping methods are not suitable for multi-mapping problems,
such as the transfer of facial attributes, art styles, or textures.
To achieve multi-mapping translation,
they need to be built for different pairs of mappings, even though some mappings share common semantics.
To overcome this limitation,
recent studies~\cite{lample2017fader,choi2017stargan,zhu2017multimodal} take both image and latent code as input to the generator to learn diverse translations.
Specifically,
the latent code can be the attribute (domain) label for multi-domain translation~\cite{lample2017fader,choi2017stargan},
or the style embedding for multi-modal translation~\cite{zhu2017multimodal}.
\begin{figure}
  \centering
  \includegraphics[width=\linewidth]{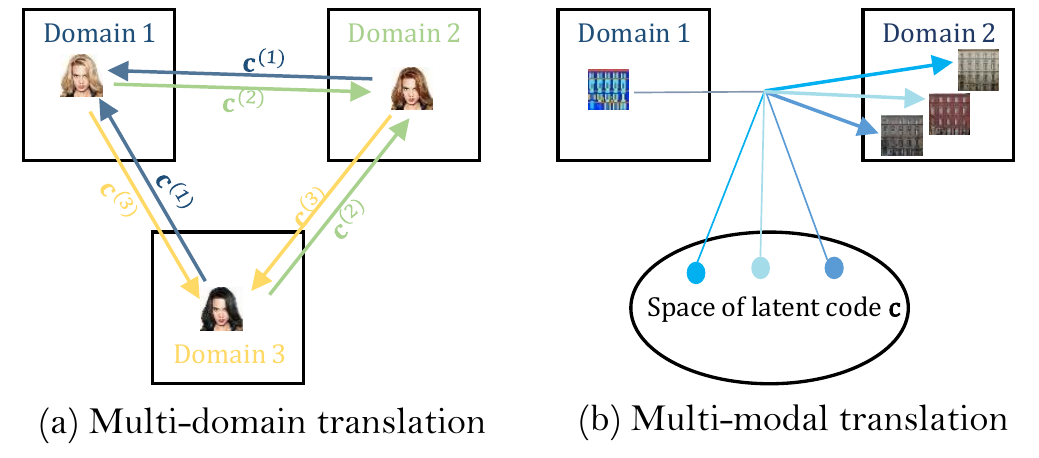}
  \caption{
   An illustration of multi-mapping indicated by the latent code $\latentcode$.
   (a) Facial attribute transfer indicated by the attribute label.
   (b) \inlinecode{label2photo} indicated by style embedding.}
  \label{fig:multi-mapping}
\end{figure}

\begin{figure*}[t]
	\centering
	\includegraphics[width=0.95\linewidth]{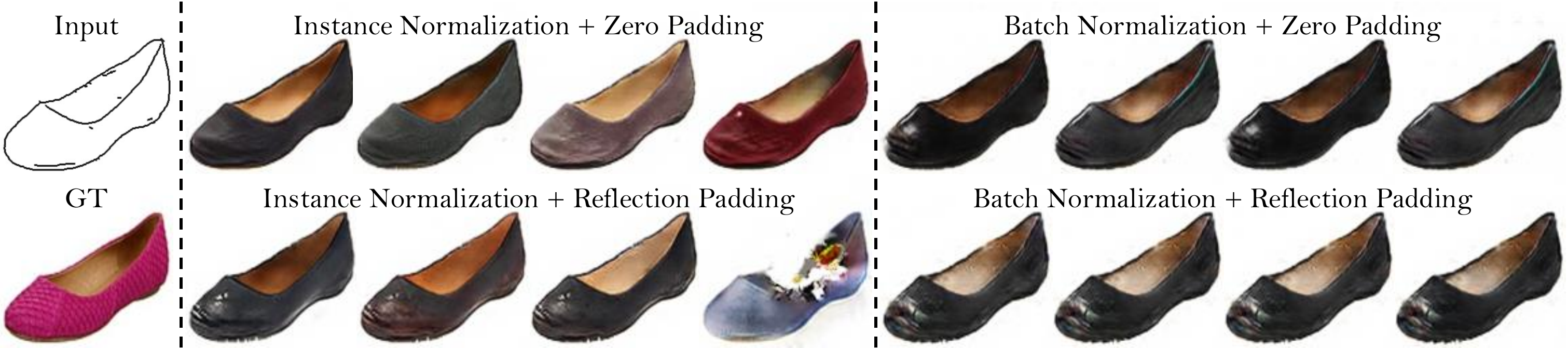}
	\caption{
		\inlinecode{Edge2photo} results sampled by BicycleGAN~\cite{zhu2017multimodal}.
		The first column shows the input image and ground truth.
		In the remaining columns,
		the  images with the same configuration are generated by sampling different latent codes.
		We can observe that the diversity and quality of BicycleGAN are sensitive to the choice of normalization and padding strategies.
}
	\label{fig:LCI_problem}
\end{figure*}

As shown in Fig.~\ref{fig:multi-mapping},
the facial attribute transfer~\cite{choi2017stargan} is a typical multi-domain translation task that aims to learn mappings among different attributes,
\eg,
black/blond/brown for hair color.
As for the multi-modal translation,
the latent code is usually sampled from a latent space with prior distribution (\eg Uniform or Gaussian priors) to indicate the cross-domain style,
such as the facade textures in the  \inlinecode{label2photo} task~\cite{zhu2017multimodal}.
Both of them attempt to capture the joint output distribution between the input image and latent code by a single generator.
But previous works~\cite{isola2017pix2pix,zhu2017multimodal} note that trivially injecting a latent code into the network did not help produce diverse results.
To prevent this mode collapse phenomenon,
recent studies focus on enforcing the generator to make use of the latent code,
such as latent regression~\cite{zhu2017multimodal} or domain classification~\cite{lample2017fader,choi2017stargan}.
However,
as illustrated in Fig.~\ref{fig:LCI_problem},
these methods are sensitive to the choice of network structure,
\eg the padding strategies or normalization operations.
To tackle this issue,
we explore the working mechanism of the multi-mapping models from the perspective of latent code injection (LCI).
Through mathematical analysis,
we show how latent code can control the target mapping by affecting the mean value of convolutional outputs.
Besides,
we find that using batch or instance normalizations in multi-mapping models results in ambiguous outputs for different mappings.
Thus the performance of the generator is sensitive to the choice of network structures.
To tackle this issue,
we introduce the \textit{consistency within diversity criteria} for multi-mapping model.
With the criteria,
we propose \textit{central biasing normalization} (CBN) as an alternative for injecting the latent code into the multi-mapping model.
The main idea of CBN is to eliminate the inconsistency of feature maps and align them according to the target mapping.
By replacing the existing LCI generator with the \textit{central biasing generator} (CBG),
we show that our method can improve the stability and performance of multi-mapping translation.
In summary, this paper makes the following contributions: 
\begin{itemize}
	\item{We show how latent code affects the mean value of the feature maps to control the target mapping in the multi-mapping model.}
	\item{We point out the potential problems of common latent code injection and propose the \textit{consistency within diversity criteria}.}
	\item{Based on the criteria, we propose central biasing normalization as an alternative to the common latent code injection strategy.}
\end{itemize}

\section{Related Work}
Benefiting from large public image repositories and high-performance computing systems,
convolutional neural networks (CNNs) have been widely used in various image processing problems in recent years.
By minimizing the loss function that evaluates the quality of results,
CNNs attempt to model the mapping between the source and target domain.
However,
it is difficult to manually design an effective and universal loss function for different tasks.
To overcome this problem,
recent studies apply generative adversarial networks (GANs) for different generation tasks because they use metric that adapts to the data rather than the task-specific evaluation.

\subsection{Image-to-Image Translation using GANs}
By staging a zero-sum game, GANs have shown impressive results in image generation~\cite{goodfellow2014generative,mao2017least,chen2016infogan,arjovsky2017wasserstein,gulrajani2017improvedwasser,radford2015dcgan}. 
The extensions of this kind of networks with conditional settings (cGAN)~\cite{mirza2014conditional} have achieved remarkable results in various conditional generation tasks such as image inpainting~\cite{pathak2016context,ren2019structureflow}, super-resolution~\cite{ledig2016photoresolution}, text2image~\cite{reed2016text2image}, facial synthesis~\cite{tian2018cr,taigman2017unsupervised,huang2018cartoon,ren2020deep}, and photo editing~\cite{brock2016neural}.
For more details of GANs,
we refer the readers to~\cite{goodfellow2016nips,creswell2018generative} for excellent overviews.

To extend cGAN as a general-purpose solution for image processing problems,
Isola~\etal~\cite{isola2017pix2pix} define the problem of image-to-image translation and propose pix2pix for tasks with data pairs.
%previous studies focused on paired~\cite{isola2017pix2pix} and unpaired translation tasks \cite{CycleGAN2017,Yi2017DualGAN,kim2017disco,liu2017UNIT}.
%Pix2pix \cite{isola2017pix2pix} uses cGANs to perform supervised learning with data pairs.
However,
many image processing tasks are ill-posed due to the lack of paired training data.
Thus, 
CycleGAN \cite{CycleGAN2017}, DiscoGAN \cite{kim2017disco}, and DualGAN \cite{Yi2017DualGAN} introduce cycle consistency to achieve unsupervised image translation.
To further regularize the unsupervised learning,
DistanceGAN~\cite{benaim2017one} proposes the distance constraints to maintain the distance between the samples before and after the mapping.
UNIT \cite{liu2017UNIT} combines variational autoencoders \cite{Kingma2013Auto} with CoGAN \cite{liu2016coupled} to learn a joint distribution of images in different domains.
These studies have promoted the development of one-to-one mapping translation,
but have shown limited scalability for multi-mapping translation.

\begin{figure*}[t]
	\centering
	{ \includegraphics[height=3.8cm]{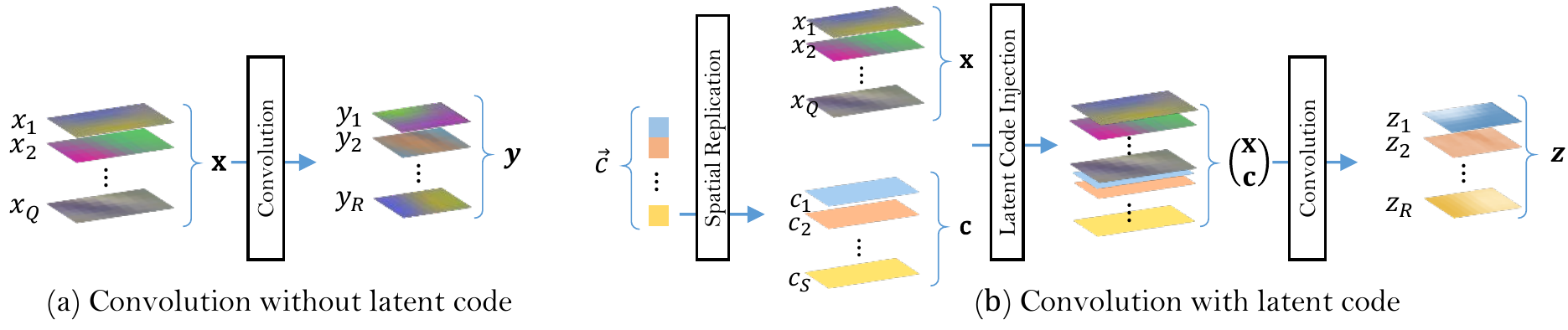} }
	\caption{Convolution operation without/with latent code.
		(a) The common convolution without latent code.
		(b) The convolution with latent code injection.
	} 
	\label{fig:pipeline}
\end{figure*}

%!TEX root = bare_jrnl.tex

\begin{table}[b]
\setlength\tabcolsep{1.6pt}
\centering
\caption{ Model comparison of image-to-image translation
}
\label{table:model_com}
\begin{tabular}{l|c|c|c|l}
	\hline
	\multicolumn{1}{c|}{Model} & \begin{tabular}[c]{@{}c@{}}Paired\\ Data\end{tabular} & \begin{tabular}[c]{@{}c@{}}Multi-\\ Domain\end{tabular} & \begin{tabular}[c]{@{}c@{}}Multi-\\ Modal\end{tabular} & \multicolumn{1}{c}{\begin{tabular}[c]{@{}c@{}}Main\\ Idea\end{tabular}} \\ \hline
	Pix2pix                    & Need                                                  & $\times$                                                & $\times$                                               & Conditional GANs                                                        \\
	CycleGAN                   & No need                                               & $\times$                                                & $\times$                                               & Cycle consistency                                                       \\
	DiscoGAN                   & No need                                               & $\times$                                                & $\times$                                               & Discover cross-domain relations                                                       \\
	DualGAN                    & No need                                               & $\times$                                                & $\times$                                               & Dual learning                                                       \\
	DistanceGAN                & No need                                               & $\times$                                                & $\times$                                               & Distance constraints                                                       \\
	UNIT                       & No need                                               & $\times$                                                & $\times$                                               & Shared latent space assumption                                          \\
	Fader Networks             & No need                                               & $\surd$                                                 & $\times$                                               & Invariant latent representation                                         \\
	StarGAN                    & No need                                               & $\surd$                                                 & $\times$                                               & Auxiliary classifier                                                    \\
	BicycleGAN                 & Need                                                  & $\times$                                                & $\surd$                                                & Bijective consistency                                                   \\ \hline
\end{tabular}
\end{table}

\subsection{Multi-Mapping Translation}
To achieve a more scalable approach for image-to-image translation, 
researchers have recently made significant progress in  multi-mapping translation~\cite{zhu2017multimodal,lample2017fader,choi2017stargan}, as compared in Table~\ref{table:model_com}.
For instance, StarGAN \cite{choi2017stargan} uses a single model and latent code to achieve multiple domain translations.
It learns multiple mappings by the auxiliary classifier~\cite{odena2016conditional}.
Fader Networks \cite{lample2017fader} learns the attribute-invariant representation for manipulating the image.
BicycleGAN \cite{zhu2017multimodal} combines VAE-GAN objects \cite{larsen2016autoencoding} and LR-GAN objects \cite{chen2016infogan,donahue2016adversarial,dumoulin2016adversarially} for a bijective mapping between the latent code and output spaces.
The common feature of these methods is that they encourage the generator to learn a joint distribution between the input image and latent code.

\subsection{Latent Code for Multi-Mapping}
For controlling multiple attributes of the generated image,
latent code~\cite{chen2016infogan,zhu2017multimodal,lample2017fader,choi2017stargan}
is introduced for targeting the salient structured semantic features. 
For instance, in the facial attribute transfer task,
the latent code $\vv{\latentcode}$ indicates the specific features,
such as gender, expression, or hair color. % \etc. 
% Injecting the latent code to the input data is a common usage of the latent code.
In existing multi-mapping models~\cite{zhu2017multimodal,lample2017fader,choi2017stargan}, 
% In this case, 
latent code $\vv{\latentcode}$ is used as an input to the convolution layer by spatial replication.
However,
this naive injection strategy is unreliable and may lead to mode collapse.
We discuss this problem in Section~\ref{sec:problems} and compare our method with StarGAN and BicycleGAN in Section~\ref{sec:exp}.

%%%%%%%%%%%%%%%%%%%%%%%%%%%%%%%%%%%%%%%%%%%%%%%%%%%%%%%%%%%%%%%%%%%%%%
\section{Common Latent Code Injection}\label{sec:rethinking}
In this section,
we first explore the existing injection mechanism by formulating the convolution operation.
Then we revisit the normalization to facilitate the later analysis in Section~\ref{sec:problems}.

\subsection{Convolution Operation without Latent Code} % Notation
Following  the notation of Convolutional Matrix Multiplication~\cite{cong2014minimizing},
we extend the matrix of numbers to the matrix of feature maps or convolution kernels.
\textit{Here, each element is a feature map or a convolution kernel instead of a number}. 

Let $x_1,x_2,\ldots,x_Q$ be
the $Q$ input feature maps (each sized $M\times N$) and $w_{r,q}$ be 
the $R\times Q$ convolution kernels (each sized $K\times L$) where $r = 1, 2, \ldots,R$ and $q =1, 2, \ldots,Q$.
Then the $R$ output feature maps $y_1,y_2, \ldots, y_R$ can be represented as
\begin{equation}
\begin{split}
y_1 =& w_{1,1} * x_1 + w_{1,2} * x_2 + \cdots + w_{1,Q} * x_Q
\\
y_2 =& w_{2,1} * x_1 + w_{2,2} * x_2 + \cdots + w_{2,Q} * x_Q
%\\
%y_3 =& w_{3,1} * x_1 + w_{3,2} * x_2 + \cdots + w_{3,Q} * x_Q
\\
&\vdots
\\
y_R =& w_{R,1} * x_1 + w_{R,2} * x_2 + \cdots + w_{R,Q} * x_Q,
\end{split}
\end{equation}
where $*$ is the convolution operation.
Further, these equations can be redefined as a special matrix/vector multiplication
\begin{equation}
	\mathbf{y} =\mathbf{W} \times \mathbf{x}, 
\end{equation}
% \begin{equation}
% \mathbf{y} = W \times \mathbf{x}
% \end{equation}
where $\mathbf{x} = (x_1, x_2, \cdots, x_Q)^T $, $\mathbf{y} = (y_1,  y_2,  \cdots,  y_R)^T $, and
\begin{equation*}
\mathbf{W} = \
\begin{pmatrix}
w_{1,1} & w_{1,2} & \cdots & w_{1,Q} \\
w_{2,1} & w_{2,2} & \cdots & w_{2,Q} \\
\vdots & \vdots & \ddots & \vdots \\
w_{R,1} & w_{R,2} & \cdots & w_{R,Q}
\end{pmatrix}.
\end{equation*}
\begin{figure*}[!ht]
	\centering
	\includegraphics[height=\ifjournal{3.8}{5}cm]{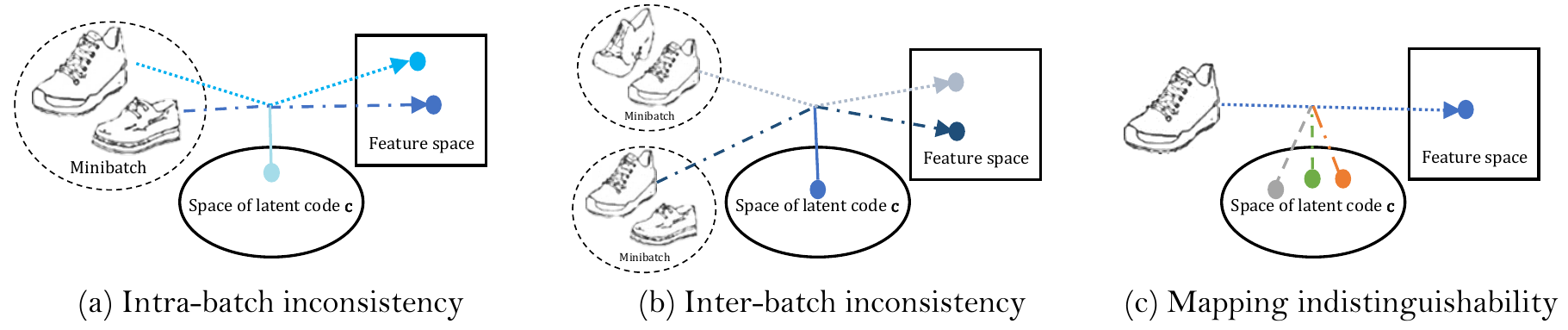}
	\caption{
		The potential problems of existing LCI models.
		(a) The intra-batch inconsistency problem of BN.
		(b) The inter-batch	inconsistency problem of BN.
		(c) The Mapping indistinguishable problem of IN.
	}
	\label{fig:problem_illu}
\end{figure*}

\subsection{Convolution Operation with Latent Code}
As shown in Fig.~\ref{fig:pipeline},
we denote the special vector $\mathbf{\latentcode}$ as the $S$ latent code feature maps,
where $\mathbf{c} = (c_1, c_2, \cdots, c_S)^T $.
The elements of $\mathbf{\latentcode}$ are replicated from the numerical element of original latent code $\vv{\latentcode}$ 
\begin{equation*}
% \vv{\latentcode}_s \quad ^{\underrightarrow{\text{ extend }}} \quad \latentcode_s,
\latentcode_s (m,n) =\vv{\latentcode}_s,
\end{equation*}
where $s =1,2,\ldots,S$; $m =1,2,\ldots,M$; $n =1, 2, \ldots,N$ and $\latentcode_s$ is a constant feature map in which every element has the same value.
We denote $v_{r,s}$ as the $R\times S$ convolution kernel that is associated with the latent code.
Then
\begin{equation}
	\mathbf{o} = \mathbf{V} \times \mathbf{\latentcode}, 
	\label{eq:v_matrix}
\end{equation}
where $\mathbf{o} = (o_1,  o_2,  \cdots,  o_R)^T $, and 
\begin{equation*}
\mathbf{V} = \
\begin{pmatrix}
v_{1,1} & v_{1,2} & \cdots & v_{1,S} \\
v_{2,1} & v_{2,2} &\cdots & v_{2,S} \\
\vdots  & \vdots  & \ddots & \vdots \\
v_{R,1} & v_{R,2} &\cdots & v_{R,S}
\end{pmatrix}.
\end{equation*}
Note each feature map $o_r$ is a constant channel as it is the linear combination of feature maps from  $\mathbf{\latentcode}$:
\begin{equation}
\label{eq:fm}
o_r = v_{r,1} * \latentcode_1 + v_{r,2} * \latentcode_2 + \cdots + v_{r,S} * \latentcode_S.
\end{equation}
% Therefore, 
The whole convolution operation from input $\mathbf{x}$ and  $\mathbf{\latentcode}$ can be represented as
\begin{align}\label{eqConstant}
\begin{autobreak}
\mathbf{z}
=(\mathbf{W},\mathbf{V})  \times \begin{pmatrix} \mathbf{x} \\ \mathbf{\latentcode} \end{pmatrix}
= \mathbf{y}  +  \mathbf{o},
\end{autobreak}
\end{align}
where $\mathbf{z} = (z_1, z_2, \cdots, z_R)^T$ is the final convolution output.

We make two observations about the convolution with latent code.
First, the final convolution output $\mathbf{z}$ can be decomposed into two separate parts  $\mathbf{y}$ and $\mathbf{o}$.
It means that the target mapping is only determined by the latent code $\mathbf{\latentcode}$.
Second,
different latent codes only provide different offsets to the outputs,
since the elements of  $\mathbf{o}$ are constant feature maps.
Thus different mappings of the same input image differ only in the mean value.
It implies that the network needs to distinguish between different mappings of an input image based on the mean value of feature maps.
Hence,
we refer to the consistency of mean value as the consistency of mapping in this work.

\subsection{Normalization}

\subsubsection{Batch Normalization}
To accelerate training and improve the performance of deep networks,
Ioffe and Szegedy introduced batch normalization (BN)~\cite{ioffe2015batch} to reduce the internal covariate shift~\cite{shimodaira2000improving} of neural networks. Although BN is designed to ease the training of discriminator,
it is also effective in deep generative models~\cite{radford2015dcgan}.
In the training stage, the input minibatch mean and standard deviation are used to normalize each feature map of the convolutional output:
\begin{equation}\label{eq4}
\text{BN}(z_r) = \frac{z_r - \Emean{Z_r} }{  \variance{Z_r} },
\end{equation}
% mini-
where $Z_r = ( z_r^{(1)}, z_r^{(2)}, \cdots ) $ is the batch of the $r$-th feature map;
$\Emean{.}$ and $\variance{.}$ are the mean operator and standard deviation operator.
During the inferencing stage, BN replaces the minibatch statistics with moving statistics.

\subsubsection{Instance Normalization}
In style transfer tasks,
Ulyanov \etal~\cite{Ulyanov2016Instance} found that critical improvement could be achieved by replacing batch normalization with instance normalization (IN).
Recent image transformation tasks~\cite{CycleGAN2017,zhu2017multimodal,choi2017stargan} confirm that IN is also useful for improving the quality of transformation results.
Unlike BN, which uses minibatch statistics,
IN only uses the statistics of the instance itself for normalization:
\begin{equation}\label{eqIN}
	\text{IN}(z_r) = \frac{z_r - \Emean{z_r} }{ \variance{z_r} } .
\end{equation}

Consequently, the feature statistics used by IN are consistent between the training and inference stages.
\subsubsection{Scale and Shift}
A simple normalization operation $\text{Norm}(.)$ for convolution output may change what it can represent.
Therefore, the affine parameters $\gamma_r,$ and $ \beta_r$ are used for restoring the representation power of the network~\cite{ioffe2015batch}:
\begin{equation}
\hat{z_r} = \gamma_r \text{Norm}(z_r) + \beta_r.
\end{equation}

\section{Analysis of Problems with Common LCI}\label{sec:problems}
According to the above analyses,
we know that the role of the latent code is to provide offsets for output feature maps.
In this section,
we further analyze the effect of latent code in the common convolution pipeline (Convolution-Normalization-Activation). 
First,
we use some examples to explain how normalization confuses the statistical mean of output feature map and degrades the model performance.
Then,
we explain why the activation function cannot help clear the ambiguity caused by the normalization.
Finally,
we propose the \textit{consistency within diversity criteria} for multi-mapping model.

\subsection{Impaired Consistency of Batch Normalization}

\subsubsection{Intra-batch Inconsistency}
Consider this minibatch $\mathbf{Z}^{(1)} = ( \mathbf{z}^{(1)}, \mathbf{z}^{(2)}, \mathbf{z}^{(3)})$,
where 
\begin{equation}
\begin{split}
\mathbf{z}^{(1)} = (\mathbf{W},\mathbf{V}) \times 
\begin{pmatrix}
\mathbf{x}^{(1)} \\ \mathbf{\latentcode}^{(1)}
\end{pmatrix}
= \mathbf{y}^{(1)} + \mathbf{o}^{(1)},
\\
\mathbf{z}^{(2)} = (\mathbf{W},\mathbf{V}) \times 
\begin{pmatrix}
\mathbf{x}^{(2)} \\ \mathbf{\latentcode}^{(1)}
\end{pmatrix}
=  \mathbf{y}^{(2)} + \mathbf{o}^{(1)},
\\
\mathbf{z}^{(3)} = (\mathbf{W},\mathbf{V}) \times 
\begin{pmatrix}
\mathbf{x}^{(3)} \\ \mathbf{\latentcode}^{(1)}
\end{pmatrix}
=  \mathbf{y}^{(3)} + \mathbf{o}^{(1)}.
\end{split}
\end{equation}
In batch $\mathbf{Z}^{(1)}$, different input instances $\mathbf{x}^{(1)}, \mathbf{x}^{(2)}, \mathbf{x}^{(3)}$ map to the same domain indicated by the latent code $\mathbf{\latentcode}^{(1)}$.
It implies that the  distribution of instances in $\mathbf{Z}^{(1)}$ should be similar in the feature space.
But without any distribution alignment operations,
the statistical mean of the feature is inconsistent in different instances
\begin{equation}
\begin{split}
&\Emean{ \text{BN}( {z^{(1)}_r} ) } - 
\Emean{ \text{BN}( {z^{(2)}_r} ) } \\
=& \Emean{ \text{BN}( {{z}^{(1)}_r} )  -  \text{BN}( {{z}^{(2)}_r} ) } \\
%=& \Emean{ \frac{{z}^{(1)}_r - 	\Emean{Z^{(1)}}}{\sqrt{ \variance{{Z^{(1)}}_r} }} - \frac{{z}^{(2)}_r - 	\Emean{Z^{(1)}}}{\sqrt{ \variance{{Z^{(1)}}_r} }}}  \\
=& \Emean{  \frac{  {{z}^{(1)}_r} - {{z}^{(2)}_r} } {  \variance{{Z^{(1)}}_r}  } }. \\
%=& \frac{\Emean{{y}^{(1)}_r - 	{y}^{(2)}_r} } {  \sqrt{ \variance{{Z^{(1)}}_r} } } .
\end{split}
\end{equation}
Since ${z}^{(1)}_r  =  {y}^{(1)}_r +{o}^{(1)}_r$ and ${z}^{(2)}_r  =  {y}^{(2)}_r +{o}^{(1)}_r$.
Hence
\begin{equation}
\begin{split}
&\Emean{ \text{BN}( {z^{(1)}_r} ) } - 
\Emean{ \text{BN}( {z^{(2)}_r} ) } \\
=& \frac{\Emean{{y}^{(1)}_r - 	{y}^{(2)}_r} } {   \variance{{Z^{(1)}}_r}  } .
\end{split}
\end{equation}
%In most case, $ \Emean{ {y}^{(1)}_r - 	{y}^{(2)}_r } \neq 0 $, so $ \Emean{ \text{BN}( {z^{(1)}_r} ) } \neq\Emean{ \text{BN}( {z^{(2)}_r} ) } $.
In most cases we have $\Emean{ {y}^{(1)}_r} \neq \Emean{ {y}^{(2)}_r }$, \textit{i.e.}, $\Emean{ {y}^{(1)}_r - 	{y}^{(2)}_r } \neq 0$, therefore 
%In most case, $ {y}^{(1)}_r$  	${y}^{(2)}_r $
\begin{equation}
\begin{split}
%&\Emean{ {y}^{(1)}_r - 	{y}^{(2)}_r } \neq 0, \\
%\Rightarrow &
\Emean{ \text{BN}( {z^{(1)}_r} ) } \neq\Emean{ \text{BN}( {z^{(2)}_r} ) }.
\end{split}
\end{equation}

Similarly, we can derive  $  \Emean{ \text{BN}( {z^{(1)}_r} ) } \neq \Emean{ \text{BN}( {z^{(2)}_r} ) } \neq \Emean{ \text{BN}( {z^{(3)}_r} ) }$.
%as illustrated in Fig.~\ref{fig:problem_illu}(a).
Although these three instances have the same mapping,
their statistical mean is inconsistent.
As the example shown in Fig.~\ref{fig:problem_illu}(a),
different input images with the same latent code cannot have a consistent mapping in the feature space because of the above problem. 
We call this phenomenon \textit{intra-batch inconsistency}.
This inconsistency is considered to be the shortcoming of BN in style transfer tasks~\cite{huang2017adain}.

\subsubsection{Inter-batch Inconsistency}

Replace $\mathbf{z}^{(3)}$ in $\mathbf{Z}^{(1)}$ with $\mathbf{z}^{(4)}$, where
\begin{equation}
\mathbf{z}^{(4)} =(\mathbf{W},\mathbf{V}) \times 
\begin{pmatrix}
\mathbf{x}^{(3)} \\ \mathbf{\latentcode}^{(2)}
\end{pmatrix}
=  \mathbf{y}^{(3)} + \mathbf{o}^{(2)}.
\end{equation}

The new batch is  $\mathbf{Z}^{(2)} = ( \mathbf{z}^{(1)}, \mathbf{z}^{(2)}, \mathbf{z}^{(4)})$.
Here, we use the statistical mean of the same mapping instances to represent the main characteristic of a specific mapping.
In batch $\mathbf{Z}^{(1)}$, the  statistical mean of the mapping indicated by $\mathbf{\latentcode}^{(1)}$ is $\Emean{\text{BN}(\mathbf{Z}^{(1)})} = 0$,
but the mean value indicated by $\mathbf{\latentcode}^{(1)}$  in batch $\mathbf{Z}^{(2)}$ is
\begin{equation}
\begin{split}
& \Emean{( \text{BN}({z^{(1)}_r}), \text{BN}({z^{(2)}_r}))} \\
= &  \frac{1}{2}\Emean{ \text{BN}( {z^{(1)}_r} ) } + \frac{1}{2}\Emean{ \text{BN}( {z^{(2)}_r} ) } \\
= &  \frac{\Emean{z^{(1)}_r}  +\Emean{z^{(2)}_r} -2\Emean{Z^{(2)}_r}}{ 2 \variance{Z^{(2)}_r} }.
\end{split}
\end{equation}
Since $ \Emean{Z^{(2)}_r}= \frac{1}{3}\Emean{z^{(1)}_r}  + \frac{1}{3}\Emean{z^{(2)}_r} + \frac{1}{3}\Emean{z^{(3)}_r}$.
Hence 
\begin{equation}
\begin{split}
& \Emean{( \text{BN}({z^{(1)}_r}), \text{BN}({z^{(2)}_r}))} \\
%= & \frac{3\Emean{Z^{(2)}_r} - \Emean{z^{(3)}_r} -2\Emean{Z^{(2)}_r}}{ 2\sqrt{ \variance{Z^{(2)}_r} }} \\
= &  \frac{\Emean{Z^{(2)}_r}  - \Emean{z^{(3)}_r}}{ 2\variance{Z^{(2)}_r} } \\
= & - \frac{1}{2} \Emean{\text{BN}(z^{(3)}_r)}.
\end{split}
\end{equation}
In most cases, $\Emean{\text{BN}(z^{(3)}_r)} \neq 0$.
Therefore, the mean value of the same mapping is always inconsistent between different batches. 
It means that the inconsistency also exists between different batches in the training stage.
We call this phenomenon \textit{inter-batch inconsistency}.
This problem causes the same input has different results in different minibatch,
as shown in Fig.~\ref{fig:problem_illu}(b).
Due to the above phenomenon,
the network has to continuously adapt to the changeful distribution of different mappings during the training stage.
Therefore, the network will suffer from covariate shift~\cite{shimodaira2000improving} which reduces its performance.
We call these problems \textit{mapping inconsistency}.

\subsection{Impaired Diversity of Instance Normalization}
The \textit{mapping inconsistency} of BN is caused by normalizing the feature statistics of the minibatch.
IN does not have this problem since it normalizes the feature statistics of an instance,
but IN will eliminate the diversity indicated by the latent code.

Consider these two instances  $\mathbf{z}^{(3)}$ and  $\mathbf{z}^{(4)}$.
With the same input  $\mathbf{x}^{(3)}$, 
different mappings are indicated by  $\mathbf{\latentcode}^{(1)}$ and $\mathbf{\latentcode}^{(2)}$.
After IN
\begin{equation}\label{eq:EliIN}
\begin{split}
&\text{IN}( z_r^{(3)} ) - \text{IN}( z_r^{(4)} )  %&= \text{IN}(z_r^{(4)} - o_r^{(4)} + o_r^{(3)} )\\
% =\frac{z_r^{(3)} - \Emean{z_r^{(3)}} }{ \sqrt{ \variance{z_r^{(3)}} }} - \frac{z_r^{(4)} - \Emean{z_r^{(4)}} }{ \sqrt{ \variance{z_r^{(4)}} }}
\\
=&\frac{y_r^{(3)} + o_r^{(1)}  - \Emean{y_r^{(3)} + o_r^{(1)}} }{  \variance{y_r^{(3)} + o_r^{(1)} }} -  
 \\& \frac{y_r^{(3)} + o_r^{(2)}  - \Emean{y_r^{(3)} + o_r^{(2)}} }{  \variance{y_r^{(3)} + o_r^{(2)}} }\\
=&\frac{y_r^{(3)}  - \Emean{y_r^{(3)} } }{  \variance{y_r^{(3)} } }  - \frac{y_r^{(3)}   - \Emean{y_r^{(3)}  } }{  \variance{y_r^{(3)} } }=0,\\
\Rightarrow & \text{IN}( z_r^{(3)} ) = \text{IN}( z_r^{(4)} ).
\end{split}
\end{equation}
Since $o_r$ is the constant feature map,
the diversity indicated by the latent code will be eliminated after IN.
%, as shown in Eq.~\ref{eq:EliIN}.
It means that the model cannot distinguish different mappings indicated by the latent code and suffer from mode collapse,
\ie,
the input image with different latent codes has same results despite the injecting information,
as shown in Fig.~\ref{fig:problem_illu}(c).
We call this problem \textit{mapping indistinguishability}.

\subsection{Effect of Activation Function}
Before outputting the normalized results,
we usually append an activation function for mapping the resulting values into the desired range.
In most cases,
the neural network requires an activation function to introduce the nonlinear property to solve the nonlinear problems~\cite{cybenko1989approximation},
and the monotonic property to ease the optimization~\cite{wu2009global}.
Although the nonlinear property will complicate the final results,
the monotonic property guarantees the mapping order after the activation function.
%the monotonic property implicates the mapping order indicated by the latent code will be preserved after the activation function.
Therefore,
the mapping problems caused by normalization still exist after activation.

\subsection{Consistency Within Diversity Criteria}
With the above potential problems,
the common LCI model has difficulty in learning the stable mapping indicated by the latent code.
Motivated by the role of latent code,
we argue that the feature statistics,
especially the mean value of the feature map,
represents the target mapping.
Therefore, we propose two criteria for designing the multi-mapping model.
To simplify the expression,
we define the multi-mapping result as $\phi(\mathbf{x}, \vv{\latentcode})$,
where $\vv{\latentcode}$ is the original latent code.
\subsubsection{Consistency criterion}
To reduce mapping inconsistency, different $\mathbf{x}^{(1)}$ and $\mathbf{x}^{(2)}$ should produce consistent outputs when given the same latent code $\vv{\latentcode}$. 
\begin{equation}\label{eq:cc}
\Emean{\phi(\mathbf{x}^{(1)}, \vv{\latentcode})} = \Emean{\phi(\mathbf{x}^{(2)}, \vv{\latentcode})}.
\end{equation}
It means that the statistical mean of the mapping function $\phi(\mathbf{x}, \vv{\latentcode})$ should not relate to the input $\mathbf{x}$.

\subsubsection{Diversity criterion}
To maintain diversity, same input $\mathbf{x}$  should produce diversity outputs when given different  $\vv{\latentcode}^{(1)}$ and  $\vv{\latentcode}^{(2)}$.
Therefore, the mean value of the outputs should not be identical: 
\begin{equation}\label{eq:dc}
\Emean{\phi(\mathbf{x}, \vv{\latentcode}^{(1)})} \neq \Emean{\phi(\mathbf{x}, \vv{\latentcode}^{(2)})}.
\end{equation}
In other words, the statistical mean of the mapping function $\phi(\mathbf{x}, \vv{\latentcode})$ should be related to the input $\vv{\latentcode}$.

Previous multi-mapping approaches~\cite{lample2017fader,zhu2017multimodal,choi2017stargan} focused on  finding effective loss function to generate diverse outputs.
As a result,
these methods meet the diversity criterion.
However,
due to the mapping inconsistency problem,
they do not satisfy the consistency criterion.
Therefore,
we refer the above criteria to as \textit{consistency within diversity criteria} to emphasize the mapping consistency.
 
\section{Proposed Method Based on Central Biasing Normalization}\label{sec:cbn}
\begin{figure}
	\centering
	\includegraphics[height=5cm]{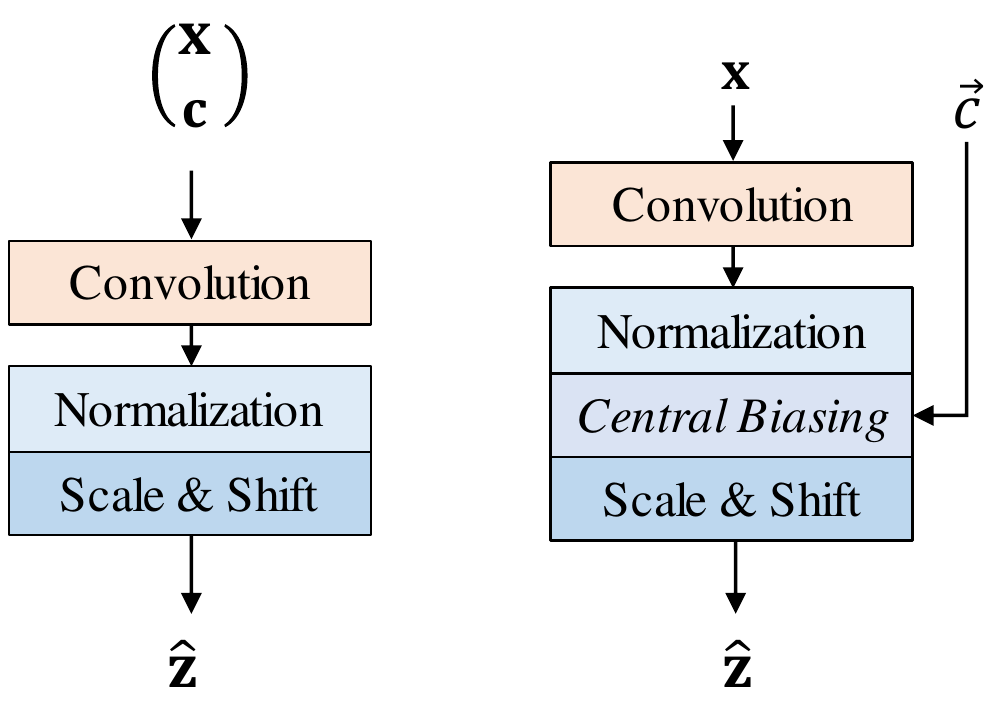}
	\caption{
		Left: a traditional convolutional block with latent code injection. Right: a convolutional block with proposed central biasing normalization.
	}
	\label{fig:sub_pipeline}
\end{figure}

In this section, we first propose the central biasing operation for common normalization according to the above criteria.
Then we apply the proposed central biasing normalization (CBN) to construct our central biasing generator (CBG) for multi-mapping translation.

\subsection{Central Biasing Normalization}
To meet the proposed criteria,
an intuitive approach is  to add related constraints or regularizations to the loss function in the training stage.
But controlling the importance of the additional loss term is a new challenge.
We can also remove the normalization to reduce the impact of potential problems.
But it causes the network to be difficult to converge.
Rethinking the potential problems discussed in Section~\ref{sec:problems},
we observe that the mapping indicated by the latent code is confusing because of the zero-centered operation in normalization.  
If we can separate the mapping learning from the feature extraction of the convolution,
we can avoid the above issues.
Therefore,
we propose a novel normalization strategy for learning the specific mapping indicated by the latent code.
Our proposed method is easy to incorporate into existing multi-mapping models and orthogonal to the ongoing exploration about learning strategy. 
\subsubsection{Formulation} \label{sec:cbn_A_1} % Definition

We first eliminate the offset of normalization feature maps to meet the consistency criterion,
which aligns the distribution center of different instances.
Then we append a bias calculated by the latent code to encourage the model to meet the diversity criterion.
We call this method as central biasing normalization.
It can be represented as
\begin{equation}
% \text{Norm}^*
\text{CBN}  ( y_r, \vv{\latentcode}) = \text{Norm}( y_r ) - \Emean{ \text{Norm}( y_r ) } + \bias_r (\vv{\latentcode}),
\label{eq:norm}
\end{equation} 
\begin{equation} \label{eq:bias}
% Y=()    =>   Y=( + )
\bias_r(\vv{\latentcode}) \defeq \tanh ( \affine(\vv{\latentcode})),
\end{equation}
where Norm(.) represents the common normalization, $\bias_r (\vv{\latentcode})$ is the bias for the $r$-th output feature map,
and $f(.)$ represents the affine transformation.
Using this initialized bias $f(\vv{\latentcode})$ also meets the \textit{design criteria},
but we argue it goes against the intention of the normalization operation.
In~\cite{ioffe2015batch},
the normalization layer is introduced to accelerate deep network training.
It can reduce the internal covariate shift of the network by ensuring the input distribution is stable for the next convolution layer.
But without any constraints,
the feature distribution  after adding bias $f(\vv{\latentcode})$  is unknown since there is no constraint on the range of bias.
To stabilize the output distribution, 
we append a \inlinecode{tanh} function to constrain the range of bias.
The final bias for the feature map is defined as Eq.~\ref{eq:bias}.
The mean value of feature is $\Emean{\text{CBN}  ( y_r, \vv{\latentcode}) } = \bias_r(\vv{\latentcode})$, where $\bias_r(\vv{\latentcode}) \in [-1,1]$.
With the scale and shift operation, the distribution after normalization is stable for the next layer.

%!TEX root = bare_jrnl.tex

\begin{table}[t]
\renewcommand\arraystretch{1.5}
\centering
\caption{The network architecture of CBG. ``$C\stimes K\stimes L$ Conv S$n$ P$m$'' denotes $C$ $n$-stride convolutional filters with ${K}\stimes L$ kernel size and $m$ sized Zero Padding or Reflection Padding. $H$ and $W$ are the height and width of the input image, respectively. 
}
\label{table:network}
\begin{tabular}{cccc}
\hline
%\multicolumn{4}{c}{\textbf{Layer}}
\multicolumn{1}{c}{\textbf{Convolution}} & \multicolumn{1}{c}{\textbf{Norm}} & \multicolumn{1}{c}{\textbf{Activation}}                       &  \multicolumn{1}{c}{\textbf{Output Size}} \\ \hline \hline                            

$64\stimes 7 \stimes 7$ Conv S1 P3 & CBN  & ReLU  & $64\stimes H\stimes W$                  \\ 
$128\stimes 4\stimes 4$ Conv S2 P1 & CBN  & ReLU & $128 \stimes \frac{H}{2} \stimes \frac{W}{2}$              \\ 
$256\stimes 4\stimes 4$ Conv S2 P1 &  CBN &  ReLU & $256 \stimes \frac{H}{4} \stimes \frac{W}{4}$               \\

$256\stimes 3\stimes 3$ Res S1 P1 & CBN & ReLU & $256 \stimes \frac{H}{4} \stimes \frac{W}{4}$               \\
$256\stimes 3\stimes 3$ Res S1 P1 & CBN & ReLU & $256 \stimes \frac{H}{4} \stimes \frac{W}{4}$               \\ 
$256\stimes 3\stimes 3$ Res S1 P1 & CBN & ReLU & $256 \stimes \frac{H}{4} \stimes \frac{W}{4}$               \\ 
$256\stimes 3\stimes 3$ Res S1 P1 & CBN & ReLU & $256 \stimes \frac{H}{4} \stimes \frac{W}{4}$               \\ 
$256\stimes 3\stimes 3$ Res S1 P1 & CBN & ReLU & $256 \stimes \frac{H}{4} \stimes \frac{W}{4}$               \\ 
$256\stimes 3\stimes 3$ Res S1 P1 & CBN & ReLU & $256 \stimes \frac{H}{4} \stimes \frac{W}{4}$               \\   \hline

$128\stimes 4\stimes 4$ TrConv S2 P1 & BN/IN & ReLU & $128 \stimes \frac{H}{2} \stimes \frac{W}{2}$               \\
$64\stimes 4\stimes 4$ TrConv S2 P1 & BN/IN & ReLU & $64 \times H \times W$              \\
$3\stimes 7\stimes 7$ TrConv S1 P3 & / & Tanh  & $3\stimes H\stimes W$                  \\  \hline

\multicolumn{1}{l}{}                          & \multicolumn{1}{l}{}     \\ 
\end{tabular}
\end{table}

\subsubsection{Proof}
By substituting Eq.~\ref{eq4} into Eq.~\ref{eq:cc},
we observe that the batch normalization does not meet the consistency criterion.
\begin{align}\label{eq:bn_discon}
\begin{autobreak}
	\Emean{ \text{BN}(z_r) } 
	= \Emean{  \frac{z_r - \Emean{Z_r} }{  \variance{Z_r} } }
	=  \frac{\Emean{z_r} - \Emean{Z_r} }{  \variance{Z_r} } 
%	= \frac{\Emean{y_r+o_r} - \Emean{Y_r + O_r} }{ \sqrt{ \variance{Z_r} }} 
	= \frac{\Emean{y_r+o_r} - \Emean{Y_r + O_r} }{  \variance{Z_r} }
	=\frac{   \Emean{y_r- \Emean{Y_r} }}{  \variance{Z_r} }  +   \frac{ \Emean{o_r- \Emean{O_r} }}{  \variance{Z_r} }.
\end{autobreak}
\end{align}
Eq.~\ref{eq:bn_discon} shows that the mean value of the feature map after BN relates to the input $\mathbf{x}$ and minibatch,
which results in \textit{mapping inconsistency}.

According to Eq.~\ref{eqIN} and Eq.~\ref{eq:dc},
we observe that instance normalization violates the diversity criterion.
\begin{align}\label{eq:in_distin}
\begin{autobreak}
	\Emean{ \text{IN}(z_r) } 
	= \Emean{  \frac{z_r - \Emean{z_r} }{  \variance{z_r} } }
	= \frac{\Emean{z_r} - \Emean{z_r} }{  \variance{z_r} } 
	= 0.
\end{autobreak}
\end{align}
Eq.~\ref{eq:in_distin} indicates that IN will eliminate the effects of the latent code,
which leads to \textit{mapping indistinguishability}. 

By contrast,
these normalizations will meet the criteria by applying the central biasing operation.
\begin{align}
\begin{autobreak}
	\Emean{  \text{CBN}(y_r, \vv{\latentcode}) } 
	= \Emean{  \text{Norm}( y_r ) - \Emean{ \text{Norm}( y_r ) } + \bias_r (\vv{\latentcode}) } 
	= \Emean{  \text{Norm}( y_r ) } - \Emean{ \Emean{ \text{Norm}( y_r ) } }  + \Emean{ \bias_r (\vv{\latentcode}) } 
	= \Emean{ \bias_r (\vv{\latentcode}) } .
\end{autobreak}
\end{align}

For specific normalizations BN and IN, CBN can be simplified as CBBN and CBIN respectively:
\begin{align}
% \text{BN} \Rightarrow
\text{CBBN} ( y_r,\vv{\latentcode}) = \frac{ y_{r} - \Emean{y_r} }{  \variance{\mathbf{Y}_r} } +  \bias_r(\vv{\latentcode}),
\\
% \text{IN} \Rightarrow
\text{CBIN} ( y_r,\vv{\latentcode}) = \frac{ y_{r} - \Emean{y_r} }{  \variance{{y}_r} } +  \bias_r(\vv{\latentcode}).
\end{align}
It should be noted that $\text{CBBN}$ eliminates the instance mean $\Emean{y_r}$ instead of batch mean $\Emean{\mathbf{Y}_r}$, and that the standard deviation of the output depends on the batch instead of instances like $\text{CBIN}$.

\subsection{Central Biasing Generator}
Instead of common LCI strategy,
we inject the latent code into the normalization layers by replacing traditional normalization with central biasing normalization,
as shown in Fig.~\ref{fig:sub_pipeline}.
We adopt the common encoder-decoder architecture~\cite{johnson2016perceptual,CycleGAN2017,choi2017stargan} 
to build the generator,
which contains two stride-2 convolution layers for downsampling, six residual blocks~\cite{he2016residual} and two stride-2 transposed convolution layers for upsampling.
The normalization layers in the downsampling and residual blocks are 
replaced with central biasing normalization layers.
We refer this generator to as the central biasing generator.
The details of the network are shown in Table~\ref{table:network}.

%!TEX root = bare_jrnl.tex
\section{Experiments}\label{sec:exp}
As discussed in Section~\ref{sec:problems}, the existing latent code injection model has some potential problems when learning multiple mappings.
Why are similar convolution pipelines in the existing works \cite{choi2017stargan,zhu2017multimodal} still working?
The reason is that these networks introduce zero padding (ZP) before the convolution operation,
which aims to control the spatial size of the output volume.
After zero padding, the latent code channel of the input volume is no longer a constant plane but has a circle of zero boundaries.
Through the convolutional operation,
the convolution output  $o_r$ in Eq.~\ref{eq:fm} is not a constant feature map.
The activation boundaries give the possibility of keeping diversity in non-boundary areas after instance normalization.
However, the problems still exist if we remove the zero padding or use other padding strategies, such as reflection padding (RP).
In this section,
we first verify the potential problems of LCI models
and then analyze the effects of padding strategies in multi-mapping models.
Finally,
we further explore the proposed central biasing normalization in several ablation studies.

\subsection{Baselines}
To verify the potential problems,
we apply different normalization (BN\&IN) and padding strategies (ZP\&RP) to the state-of-the-art multi-mapping translation models.
\subsubsection{StarGAN}
To model multi-domain mappings with a single model, StarGAN introduces the auxiliary classier~\cite{odena2016conditional} for the discriminator.
Since StarGAN uses the attribute label as the latent code $\vv{\latentcode}$,
the latent space is discrete, and the state of $\vv{\latentcode}$ is limited.
\subsubsection{BicycleGAN}
To model the distribution of possible outputs,
BicycleGAN combines the VAE-GAN~\cite{larsen2016autoencoding} and LR-GAN~\cite{chen2016infogan,donahue2016adversarial,dumoulin2016adversarially} objectives for encouraging a bijective mapping between the latent and output spaces.
Since BicycleGAN uses the encoder with a Gaussian assumption to extract the latent code $\vv{\latentcode}$,
the latent space is continuous and the state of $\vv{\latentcode}$ is varied.
To further compare the effect between LCI and CBN,
we also remove the encoder of BicycleGAN and use the noise vector as the latent code.
Without the incentive for the generator to make use of the latent code,
BicycleGAN degenerates into pix2pix~\cite{isola2017pix2pix}.

\subsection{Datasets}
To evaluate the performance of StarGAN,
we study the facial attribute transfer on the \href{http://mmlab.ie.cuhk.edu.hk/projects/CelebA.html}{CelebA}\footnote{\href{http://mmlab.ie.cuhk.edu.hk/projects/CelebA.html}{http://mmlab.ie.cuhk.edu.hk/projects/CelebA.html}} dataset~\cite{liu2015deep}.
According to~\cite{choi2017stargan},
we select 2,000 test images from 202,599 face images and retain the rest to training.
Seven facial attributes are selected for experimentations:
hair color (black, blond, brown), gender (male/female), and age (young/old).
Besides,
we compare the performance of BicycleGAN-based models on several multi-model translation tasks, 
including \href{http://people.csail.mit.edu/junyanz/projects/gvm}{edge2photo}\footnote{\href{http://people.csail.mit.edu/junyanz/projects/gvm}{http://people.csail.mit.edu/junyanz/projects/gvm}}~\cite{yu2014fine,zhu2016generative} and \href{http://cmp.felk.cvut.cz/~tylecr1/facade}{label2photo}\footnote{\href{http://cmp.felk.cvut.cz/~tylecr1/facade}{http://cmp.felk.cvut.cz/~tylecr1/facade}}~\cite{tylevcek2013spatial}.
The default training and test sets of these datasets are used for all experiments.
For each task,
we resize the image to 128$\times$128 resolution and follow the suggested training strategy in ~\cite{choi2017stargan,zhu2017multimodal}.

\begin{figure*}[!tb]
	\centering
	\includegraphics[height=\ifjournal{17.05}{8}cm]{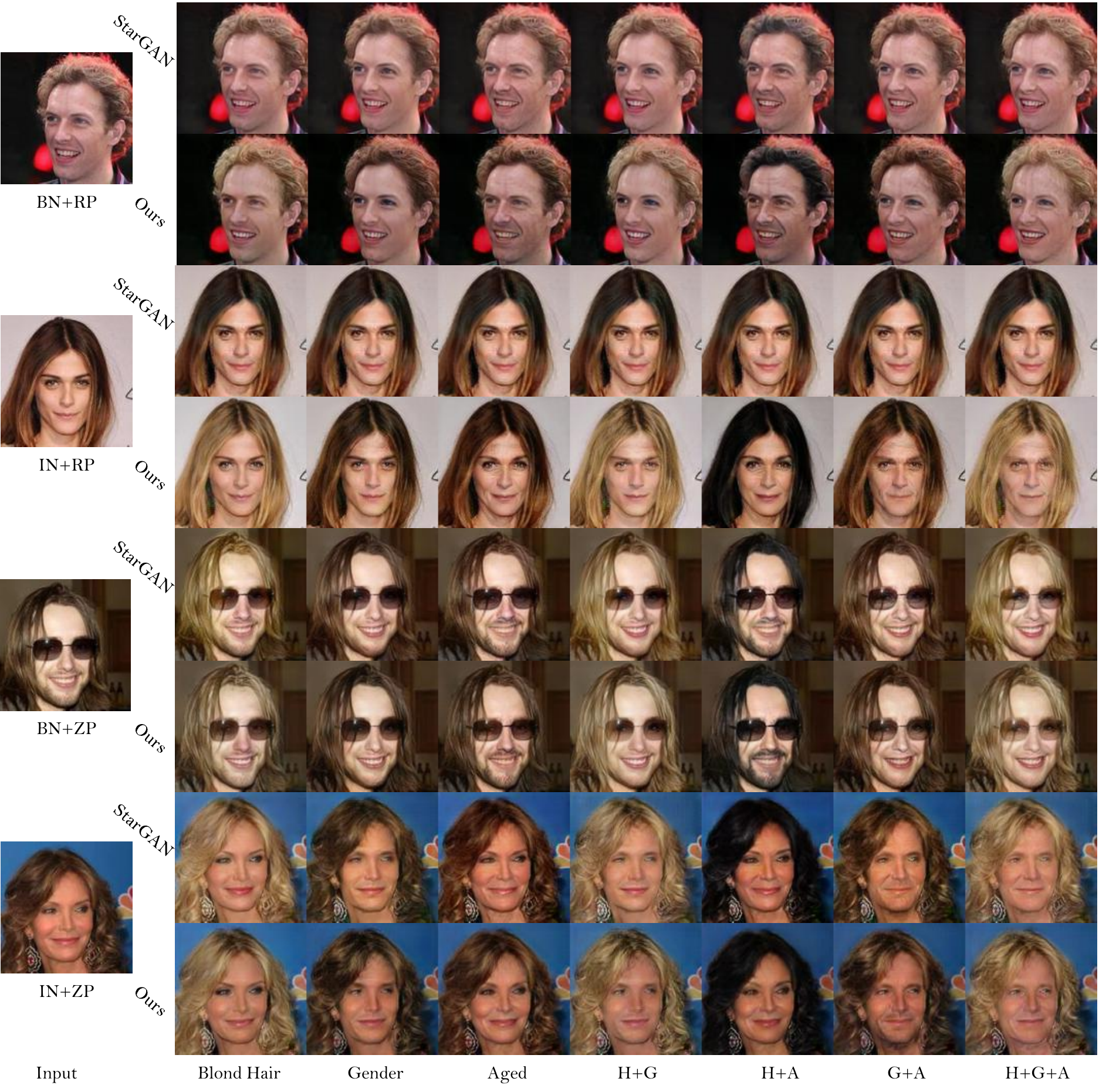}
	\caption{Facial attribute transfer results. The first column shows the input images, the next three columns show the single attribute transformation results, and the rest of the columns show the multi-attribute transfer results. H: Hair color, G: Gender, A: Aged.
	}
	\label{fig:faceedit}
\end{figure*}

\subsection{Metrics}
\subsubsection{Classification Accuracy}
To judge the translation accuracy of the multi-domain translation,
%To judge the specific domains image translation capabilities,
we compare the classification accuracy of facial attributes on synthesized images.
Firstly, we train a ResNet-18~\cite{he2016residual} as the multi-label classifier on the \inlinecode{CelebA} dataset.
Then we translate each test image to all possible targets (12 domain of facial attributes).
Finally,
we classify the translated images using the ResNet classifier.

\subsubsection{Diversity and Consistency Score}
To compare the diversity of different multi-modal models, 
we compute the averaged perceptual distance in feature space.
As suggested in~\cite{zhu2017multimodal},
we use the cosine similarity (CosSim) to evaluate the feature distance of the VGG-16 network~\cite{Simonyan2014Very} pre-trained on ImageNet.
We average across spatial dimensions and sum across the five convolution layers preceding the pool layers.
As in Zhu \etal~\cite{zhu2017multimodal}, 
the perceptual distance is defined as
($5.0-\sum_{i=1}^5 \text{CosSim}_i$ ).
The larger the perceptual distance, the greater the difference between the two images.
To evaluate the diversity of different models,
we randomly sample an input image and use a pair of random latent code to generate images.
Then, we compute the average distance between 2,000 pairs of generated images.
In addition to the diversity,
the mapping consistency is also important to the multi-modal translation.
\begin{table}[htb]
\centering
\caption{The classification accuracy for facial attribute transfer.}
\label{table:star}
\begin{tabular}{cl|ccc|c}\hline
	\multicolumn{2}{c|}{Method}      & \multicolumn{1}{l}{Hair color} & \multicolumn{1}{l}{Gender} & \multicolumn{1}{l|}{Aged} & \multicolumn{1}{l}{Average} \\ \hline
	\multirow{2}{*}{BN+RP} & StarGAN & 70.90\%                          & 61.16\%                      & 61.57\%                     & 64.54\%                       \\
	& Ours     & \textbf{81.93\%}                 & \textbf{68.00\%}             & \textbf{63.21\%}            & \textbf{71.05\%}              \\ \hline
	\multirow{2}{*}{IN+RP} & StarGAN & 33.33\%                          & 50.00\%                      & 50.00\%                     & 44.44\%                       \\
	& Ours     & \textbf{96.46\%}                 & \textbf{88.55\%}             & \textbf{86.05\%}            & \textbf{90.36\%}              \\ \hline
	\multirow{2}{*}{BN+ZP} & StarGAN & 84.25\%                          & \textbf{79.53\%}             & 66.09\%                     & 76.62\%                       \\
	& Ours     & \textbf{88.63\%}                 & 76.32\%                      & \textbf{67.76\%}            & \textbf{77.57\%}              \\ \hline
	\multirow{2}{*}{IN+ZP} & StarGAN & 95.25\%                          & 88.38\%                      & 86.05\%                     & 89.89\%                       \\
	& Ours     & \textbf{95.74\%}                 & \textbf{90.20\%}             & \textbf{86.11\%}            & \textbf{90.68\%}              \\ \hline				
	\multicolumn{2}{c|}{Real images}  & 		 91.56\%&      97.25\%&    88.74\%& 92.52\%   \\ \hline
\end{tabular}
\end{table}
\begin{table}[htb]
\centering
\caption{The Fréchet Inception Distance and human evaluation score for facial attribute transfer.}
\label{table:fid_human}
\begin{tabular}{cl|cc}\hline
	\multicolumn{2}{c|}{Method}      & \multicolumn{1}{l}{FID} & \multicolumn{1}{l}{Human evaluation}  \\ \hline
	\multirow{2}{*}{BN+RP} & StarGAN & 20.71                          & 34.67\%                                          \\
	& Ours     & \textbf{19.83}                 & \textbf{65.33\%}                         \\ \hline
	\multirow{2}{*}{IN+RP} & StarGAN & 20.26                          & 15.63\%                                           \\
	& Ours     & \textbf{16.50}                 & \textbf{84.37\%}                           \\ \hline
	\multirow{2}{*}{BN+ZP} & StarGAN & 17.53                          & 48.20\%                                 \\
	& Ours     & \textbf{17.45}                 & \textbf{51.80\%}                                  \\ \hline
	\multirow{2}{*}{IN+ZP} & StarGAN & 16.72                          & 49.67\%                                            \\
	& Ours     & \textbf{16.62}                 & \textbf{50.33\%}                         \\ \hline		
\end{tabular}
\end{table}
\begin{figure*}[ht!]
	\centering
	\includegraphics[height=8.5cm]{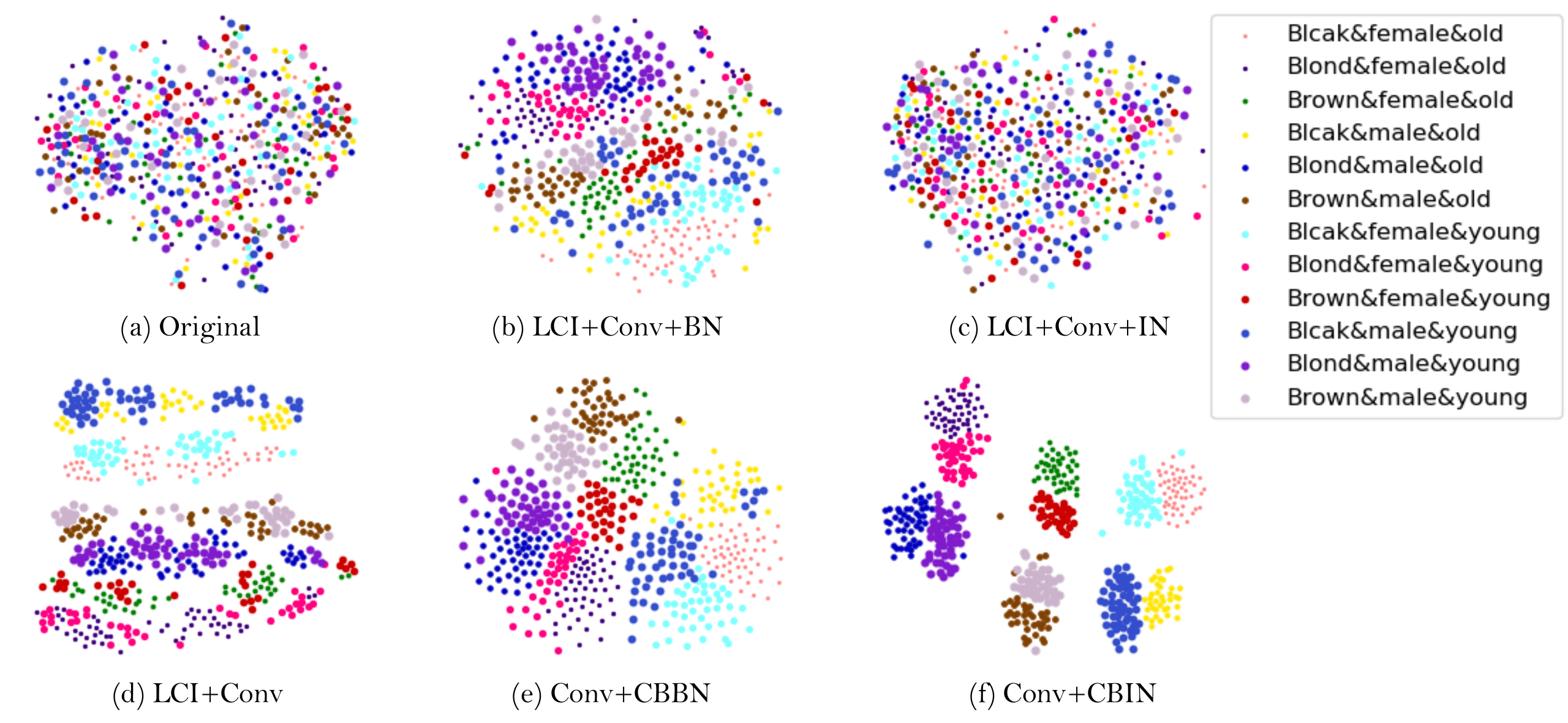}
	\caption{
		Feature visualization of facial attribute transfer.
		Different colors represent different mappings indicated by the latent code.
	}
	\label{fig:visual}
\end{figure*}
We first use the latent code encoded by ground truth to indicate the image generation.
Then we calculate the average reconstruction loss to measure the consistency of models.
More specifically,
the consistency score is calculated by ($1.0- ||G(Enc(y),x) - y||_1$),
where $x$ and $y$ are the input and target images, $Enc$  is the encoder, and $G$ represents the generator. 
Intuitively,
the higher the consistency score,
the stronger the ability of the model to generate images with the specified style.

\subsubsection{Fréchet Inception Distance}
To measure the quality of the generated images,
we adopt Fréchet Inception Distance (FID)~\cite{heusel2017gans} to evaluate the similarity between the generated images and the real images.
Specifically, 
we translate each facial image to all possible combinations of attributes in the multi-domain translation task,
and randomly generate 10 different samples for each image in the multi-modal translation task.
Then we compute the FID score between the distribution of generated images and real images.
The lower the FID score, the better the quality of the generated images.

\subsubsection{Human Perception}
To compare the realism of translation outputs,
we conduct a human perceptual study on Amazon Mechanical Turk (AMT).
In the multi-domain translation task,
the workers are given a description of attributes and two translation outputs from the methods with and without CBN.
They are given unlimited time to select which image looks more realistic and fits the given attributes.
Similarly,
in the multi-modal translation task,
the workers are required to choose a more realistic one from the images generated by different methods.
We randomly generate 500 questions for each comparison,
and each question is finished by 5 different workers.
A higher score indicates that the generated images are more realistic.

\subsection{Verification of Potential Problems}
\subsubsection{Experiments on CelebA Dataset}
The qualitative comparison results are shown in Fig.~\ref{fig:faceedit}.
We observe that StarGAN fails to generate diversity outputs when it is equipped with reflection padding.
These results have improved after it is equipped with zero padding.
By replacing the original generator of StarGAN with the proposed CBG model,
we observe that the results in all settings are both diverse and realistic.

The quantitative results further confirm the observations above.
As shown in Table~\ref{table:star},
we observe that the classification accuracy of the generated results is improved after StarGAN is equipped with CBN.
The FID scores and human perception results,
shown in Table~\ref{table:fid_human},
also suggest that the proposed CBN can help the model to learn multiple mappings more efficiently and to improve the quality of the generated images.

In order to illustrate the potential problems of the LCI model more intuitively,
we visualize the feature embeddings by using t-SNE~\cite{maaten2008visualizing} in Fig.~\ref{fig:visual}.
All of the models are equipped with reflection padding to satisfy the constant assumption in Eq.~\ref{eq:fm}.
Since each input image is randomly translated to a possible target domain,
we can observe that the t-SNE embeddings of original inputs are chaotic in Fig.~\ref{fig:visual}(a).
Despite the injection of latent code,
the embeddings cannot be well clustered,
as shown in Fig.~\ref{fig:visual}(d).
This observation also holds after batch normalization because of the mapping inconsistency problem.
After instance normalization,
the embeddings are chaotic again because of the mapping indistinguishable problem.
In contrast,
the embeddings of different mappings are clustered well after central biasing normalization.

\begin{figure*}[htb]
	\centering
	\includegraphics[height=\ifjournal{17.5}{6.5}cm]{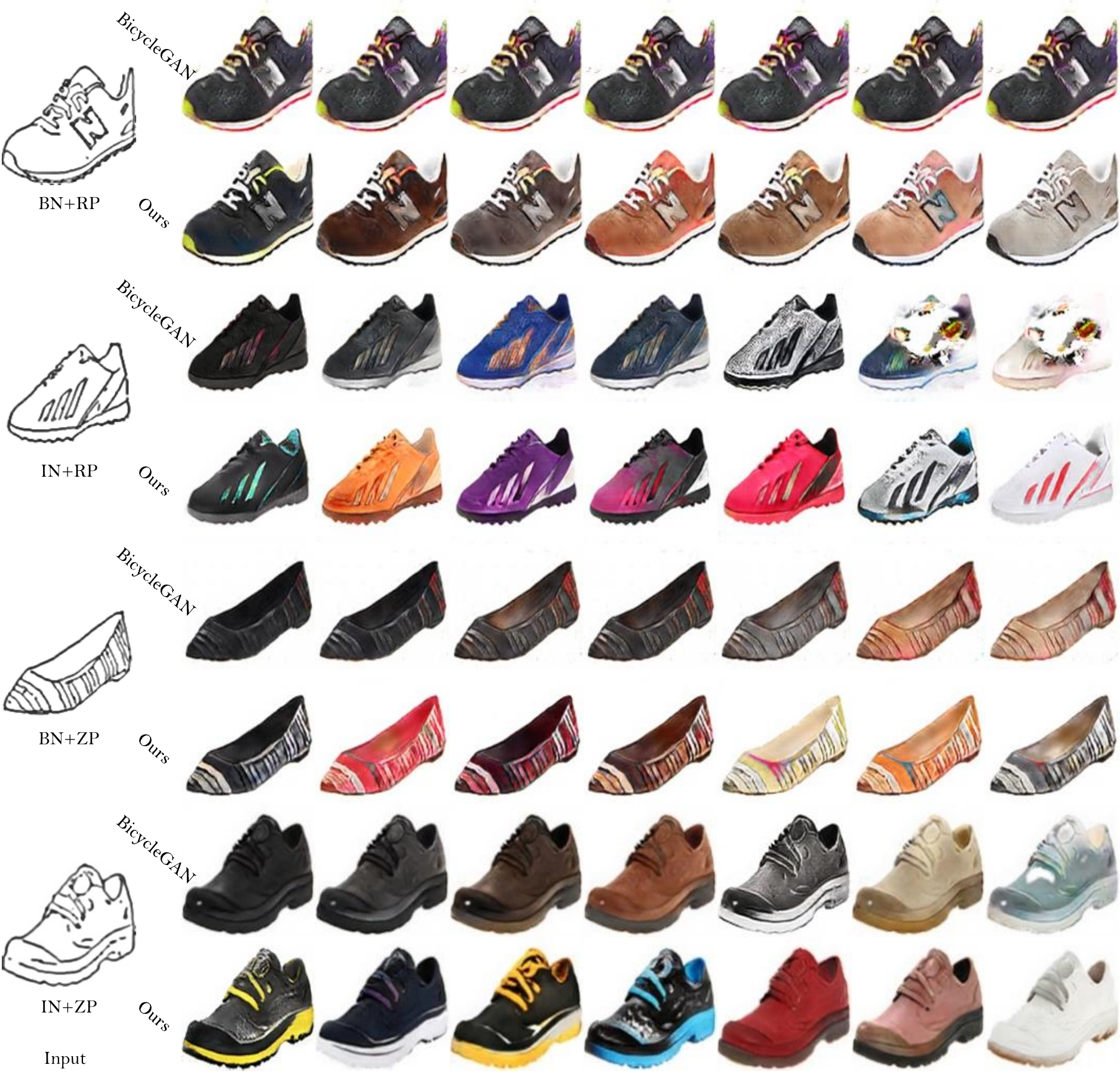}
	\caption{\inlinecode{Edge2photo} results.
		The first column shows the input images and  the remaining columns are the transfer results  indicated by random latent codes.
	}
	\label{fig:sketch2photo}
\end{figure*}

%!TEX root = bare_jrnl.tex
\begin{table}[!htb]
\centering
\caption{The diversity and consistency scores for \inlinecode{edge2photo} task. 
}
\label{table:bicycle_div_cons}
\begin{tabular}{ll|cc}
	\hline
	\multicolumn{2}{c|}{Method}         & \multicolumn{1}{l}{Diversity} & \multicolumn{1}{l}{Consistency} \\ \hline
	\multirow{2}{*}{BN+RP} & BicycleGAN & 0.069                                        & 0.906                                                               \\
	& Ours        & \textbf{0.867}                               & \textbf{0.951}                                              \\ \hline
	\multirow{2}{*}{IN+RP} & BicycleGAN & 1.025                                        & 0.949                                                               \\
	& Ours        & \textbf{1.652}                               & \textbf{0.966}                                           \\ \hline
	\multirow{2}{*}{BN+ZP} & BicycleGAN & 0.492                                        & 0.939                                                               \\
	& Ours        & \textbf{0.754}                               & \textbf{0.951}                                             \\ \hline
	\multirow{2}{*}{IN+ZP} & BicycleGAN & 1.106                                        & 0.952                                                            \\
	& Ours        & \textbf{1.637}                               & \textbf{0.966}                                               \\ \hline
	\multicolumn{2}{c|}{Real images}  & 		3.481&      N/A   \\ \hline
\end{tabular}
\end{table}
%!TEX root = bare_jrnl.tex
\begin{table}[!htb]
	\centering
	\caption{The Fréchet Inception Distance and human evaluation score for \inlinecode{edge2photo} task. 
	}
	\label{table:bicycle_FID_human}
	\begin{tabular}{ll|cc}
		\hline
		\multicolumn{2}{c|}{Method}         & \multicolumn{1}{l}{FID} & \multicolumn{1}{l}{Human evaluation} \\ \hline
		\multirow{2}{*}{BN+RP} & BicycleGAN & 85.67                                        & 43.43\%                                                               \\
		& Ours        & \textbf{41.64}                               & \textbf{56.57\%}                                              \\ \hline
		\multirow{2}{*}{IN+RP} & BicycleGAN & 57.29                                        & 45.27\%                                                               \\
		& Ours        & \textbf{34.23}                               & \textbf{54.73\%}                                           \\ \hline
		\multirow{2}{*}{BN+ZP} & BicycleGAN & 68.18                                        & 45.93\%                                                               \\
		& Ours        & \textbf{42.32}                               & \textbf{54.07\%}                                             \\ \hline
		\multirow{2}{*}{IN+ZP} & BicycleGAN & 44.30                                        & 49.10\%                                                            \\
		& Ours        & \textbf{32.19}                               & \textbf{50.90\%}                                               \\ \hline
	\end{tabular}
\end{table}

\subsubsection{Experiments on Edge2photo Dataset}
To verify the potential problems of LCI model in multi-modal translation task,
we evaluate BicycleGAN by the edge2photo task.
As the qualitative comparison shows in Fig.~\ref{fig:sketch2photo},
we can get observations similar to the facial attribute transfer above:
the CBN based models produce more realistic results while maintaining diversity.
As shown in Table~\ref{table:bicycle_div_cons},
we can observe that our method obtains higher diversity and consistency scores than BicycleGAN under the same settings.
The FID scores and human perception results shown in Table~\ref{table:bicycle_FID_human} further demonstrate our analysis of the potential problems.

\begin{figure}[t!]
	\centering
	\includegraphics[height=5cm]{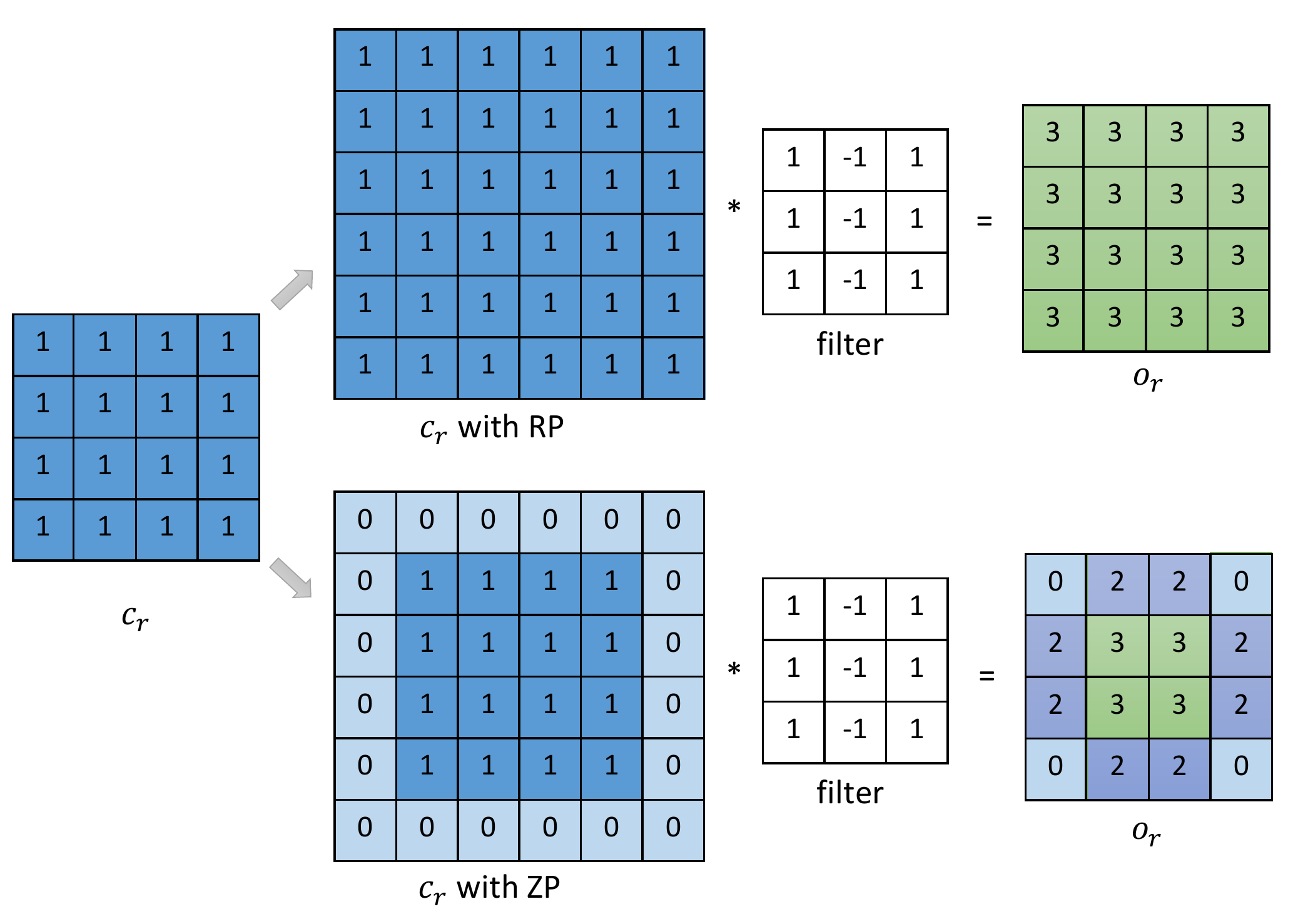}
	\caption{
		Different padding operations. The first column is the source latent code channel $\latentcode_r$. The first row in the rest of the columns is the convolution with reflection padding, and the second row is the convolution with zero padding.
	}
	\label{fig:padding}
\end{figure}%

\subsection{Analysis of Padding Strategies}
In the above experiments,
we observe that the improvements of the model with ZP are smaller than the model with RP after applying CBN,
especially in facial attribute transfer task.
To analyze this phenomenon,
let us revisit the padding strategy through Fig.~\ref{fig:padding}.
As discussed in the previous sections,
the latent code just provides a constant offset in RP mode,
so there are mapping inconsistency and mapping elimination problems.
For ZP mode, the convolution of latent code  provides the  feature map which contains non-constant boundaries and a constant center region.
Therefore, the network has the ability to control the distribution of features for different mappings  after normalization.
Since the mean value of the feature is normalized to zero after IN,
the common LCI model with ZP is inconsistent with our \textit{diversity criteria}.
But apart from the non-constant boundaries,
the input  latent code does provide the constant offsets for different output feature maps.
We think the key to distinguishing different mappings is also the mean value in non-boundary feature regions.

\begin{figure}[t!]
	\centering
	\includegraphics[height=5cm]{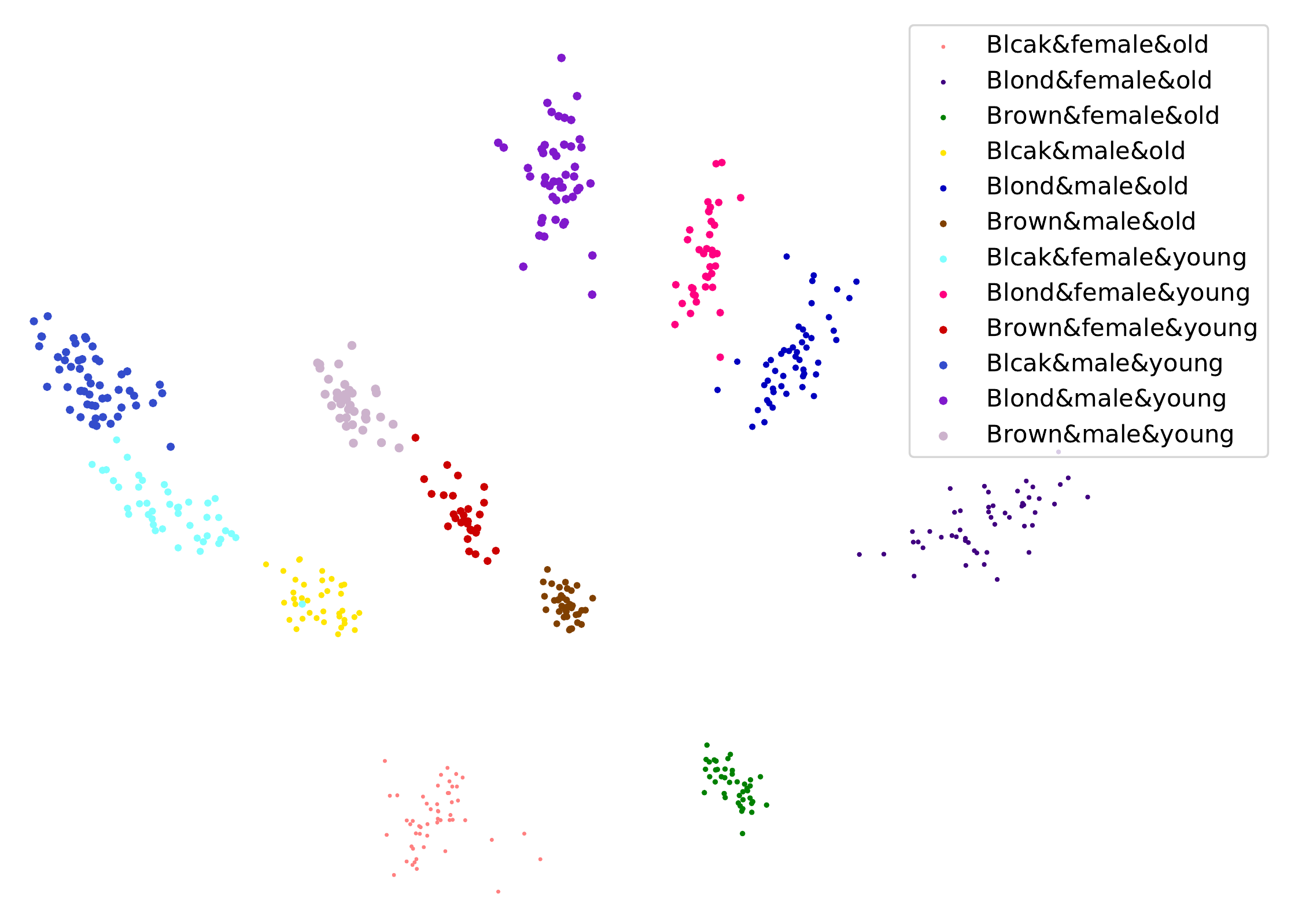}
	\caption{
		The results of k-means clustering from 2-dimensional statistical data.
		Different colors represent different mappings indicated by the latent code.
	}
	\label{fig:clu}
\end{figure}

To verify the above conjecture,
we test the StarGAN under the settings of IN+ZP and find that the statistics of non-boundary feature regions are strongly related to the target mapping.
We first sample the feature maps $\mathbf{z}$ which are the normalized outputs after convolution operation.
Then,
we calculate the mean value of features without considering the boundary area affected by padding operation.
To reduce the computational complexity,
we use principal component analysis (PCA) to reduce the dimension of the statistical embeddings.
While retaining $96\%$ variance,
we reduce the data from 64 to 6 dimensions.
After performing k-means (k=12) clustering in the low dimensional data,
we find that each cluster represents a kind of mapping,
as shown in Fig.~\ref{fig:clu}.
%We further reduce the data dimension for visualization,
According to the above experiment,
we confirm that ZP can alleviate the mapping problems discussed above.
Therefore,
when the model is equipped with ZP,
the improvement of our method is not obvious in the facial attribute transfer task.
But unlike CBN, which can align the statistical mean of the same mapping,
we observe the clusters are not compact in Fig.~\ref{fig:clu}.
It implies the decision boundaries of different mappings are not sharp.
When there are the countless target mappings,
\eg edge2photo task,
the models without CBN suffer from mapping inconsistency again and produce monotonous outputs.

\subsection{Ablation Studies and Analyses}
%!TEX root = bare_jrnl.tex
\begin{table}[!tb]

\centering
\caption{The perceptual distance for CBG with different constraint functions. 
}
\label{table:ati_div}
\begin{tabular}{l|cc|c}
	\hline
	Constraint     & edge$\rightarrow$photo & label$\rightarrow$photo & Average        \\ \hline
	None    &  ~~1.577$^{\rm a}$                  & \textbf{1.609}          & 1.593          \\
	Tanh    & \textbf{1.637}         & 1.562                   & \textbf{1.600} \\
	Sigmoid & 1.427                  & 1.166                   & 1.297          \\ \hline
\end{tabular}
\begin{threeparttable}
\begin{tablenotes}
	\scriptsize \item[a] The results were calculated from the generated images before mode collapse.
\end{tablenotes}
\end{threeparttable}
\end{table}
\begin{figure*}[!ht]
	\centering
	\includegraphics[width=\linewidth]{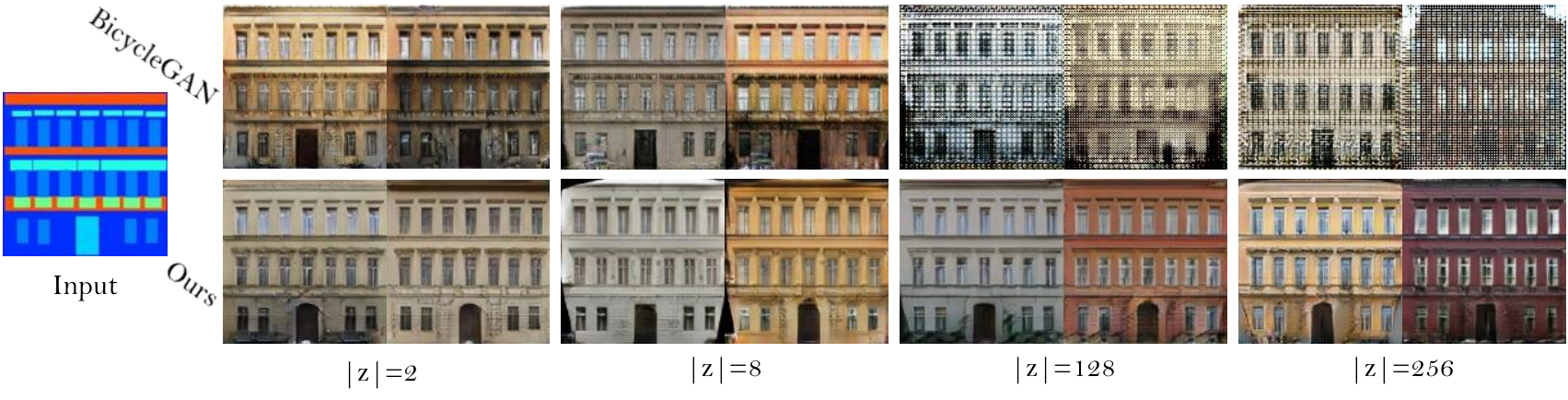}
	\caption{
		\inlinecode{Label2photo} results with varying length of the latent code.
		The images in each pair show randomly generated samples.
		We observe that larger $|z|$ can encode more information for expressive results but is not conducive to BicycleGAN output densely fill results.
	}
	\label{fig:lenghth_code}
\end{figure*}
\begin{figure*}[!ht]
	\centering
	\includegraphics[width=\linewidth]{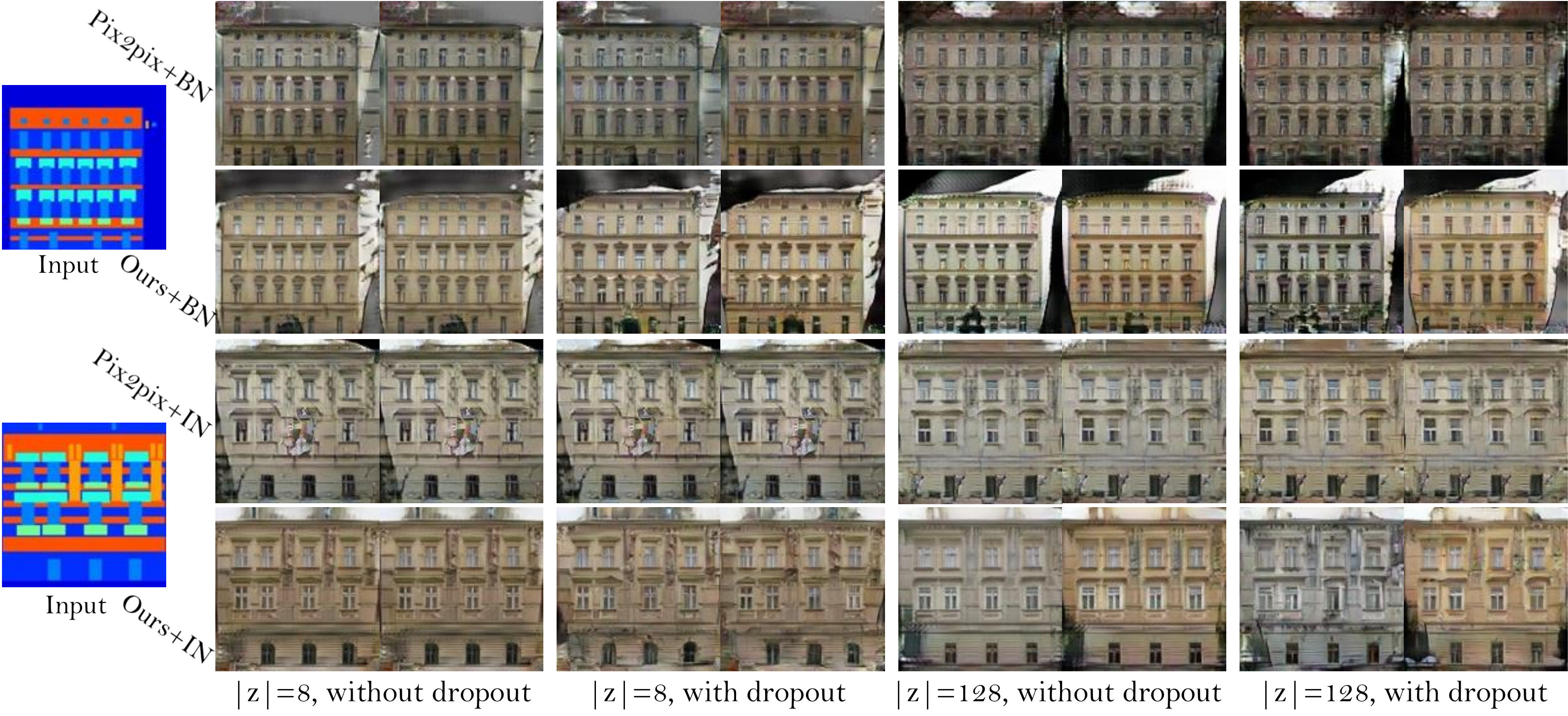}
	\caption{
		Qualitative comparison of pix2pix based models.
		The images in each pair are generated by sampling different latent codes.
	}
	\label{fig:pix2pix}
\end{figure*}

\subsubsection{Range of Bias}
As we analyzed in Section~\ref{sec:cbn_A_1},
adding the unconstrained bias goes against the intention of the normalization.
Therefore,
it is necessary to constrain the range of bias.
Here,
we further explore the impact of different bias ranges by BicycleGAN.
We extend the range of bias by removing  the constraint function
and narrow the range by replacing \inlinecode{tanh} with \inlinecode{sigmoid}.
As shown in Table~\ref{table:ati_div},
we find that \inlinecode{sigmoid} is inferior to \inlinecode{tanh}  in diversity performance.
It implies that the narrow range of bias will limit the representation of the network.
This finding is further validated when we remove the constraint function to expand the range in the label$\rightarrow$photo task.
But we also observe that the model without constraint function is unstable in the training stage because the feature distribution is unbounded.
This phenomenon is obvious in the edge$\rightarrow$photo task due to its diverse target styles.
In this task,
we find that the generator is easy to collapse when we remove the constraint function.
Therefore, we propose to apply the \inlinecode{tanh} function to constrain the range of bias while maintaining the capacity of the network.

\begin{table}[!tb]
\centering
\caption{ Generator parameters with different length of latent code.
}
\label{table:parm}
\setlength{\tabcolsep}{7mm}
\renewcommand\arraystretch{1.2}
\begin{tabular}{lcc}
	\hline
	\multirow{2}{*}{Base model} & BicycleGAN & CBG             \\ \cline{2-3} 
	& 39.9M      & \textbf{8M}     \\ \hline
	$|c|=2$                       & +78K       & \textbf{+6.9K}  \\
	$|c|=8$                       & +312K      & \textbf{+27.5K} \\
	$|c|=128$                     & +4.9M      & \textbf{+440K}  \\
	$|c|=256$                     & +9.75M     & \textbf{+880K}  \\ \hline
\end{tabular}
\end{table}

\subsubsection{The Length of Latent Code}
As in Zhu~\etal~\cite{zhu2017multimodal}, we explore the effect of latent code length on model performance.
Under varying numbers of dimensions of latent codes $\left\{2, 8, 128, 256\right\}$,
we test the default BicycleGAN, which uses IN and ZP, and CBG with same settings.
Similar to the results of  Zhu~\etal~\cite{zhu2017multimodal},
a high-dimensional latent code can potentially encode more information for image generation at the cost of making sampling quite difficult for the common LCI model.
In contrast,
our method shows stable performance when the dimension of the latent code is high enough, as shown in Fig.~\ref{fig:lenghth_code}.
Besides,
the parameters introduced by CBN are typically negligible.
Table~\ref{table:parm} shows the comparison of parameters between BicycleGAN and CBG.
Due to the replication of latent code in the common LCI model,
the convolutional parameters for latent code (convolution matrix $\mathbf{V}$ in Eq.~\ref{eq:v_matrix}) are redundant.
For a fair comparison, we just use the suggested latent code with dimension $8$~\cite{zhu2017multimodal}  in the above experiments.

\subsubsection{The Incentive of Training}
\begin{figure}[t]
	\centering
	\includegraphics[width=\linewidth]{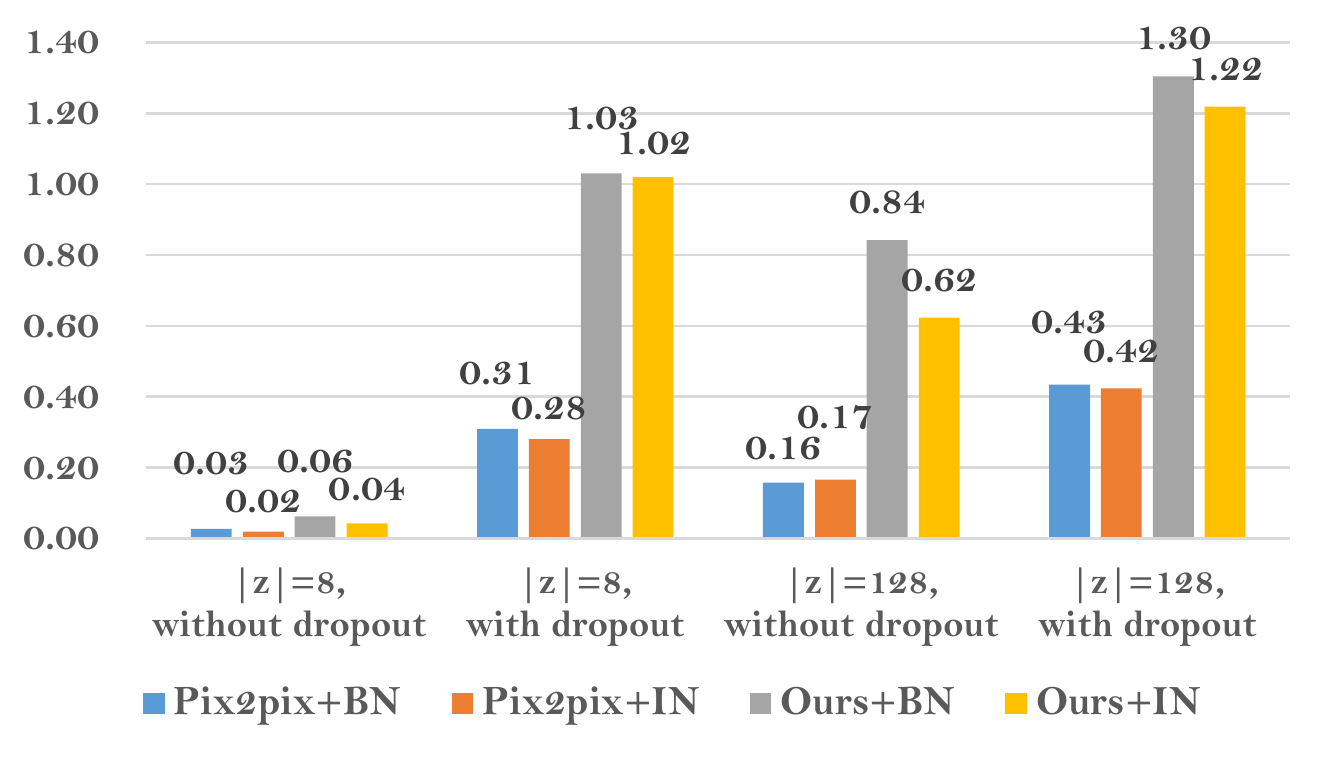}
	\caption{
		The perceptual distance of pix2pix based models.
		We compare diversity results on the \inlinecode{label2photo} across different construction strategies. 
	}
	\label{fig:pix2pix_div}
\end{figure}

To further explore the effectiveness of CBN when it lacks the incentive to make use of the latent code,
we randomly sample Gaussian noise as the latent code injects into pix2pix~\cite{isola2017pix2pix},
which can be considered as the degradation of BicyeleGAN.
Similar to~\cite{isola2017pix2pix},
we apply dropout with a rate of $50\%$ to the generator (Convolution-Norm-Dropout-ReLU) to increase the stochasticity in the output.
We also extend the dimension of the latent code to reinforce its effectiveness.
The diversity results are shown in Fig.~\ref{fig:pix2pix_div},
and the qualitative comparison results are presented in Fig.~\ref{fig:pix2pix}.
Similar to the results in~\cite{isola2017pix2pix,zhu2017multimodal},
we observe that injecting noise into the original pix2pix does not produce a large variation.
The situation has not improved even though we applied dropout and extended the dimension of the latent code.
The same results also appear when we directly replace the generator in pix2pix with CBG.
After applying the dropout operation,
we observe that the diversity score is improved by introducing stochasticity in the image structure.
By extending the dimension of latent code,
the outputs become diverse and the score is further improved.
When we apply both strategies to CBG,
we get the best diversity performance in the pix2pix based model.

\subsubsection{Convergence Rate}
\begin{figure}[t]
	\centering
	\includegraphics[width=\linewidth]{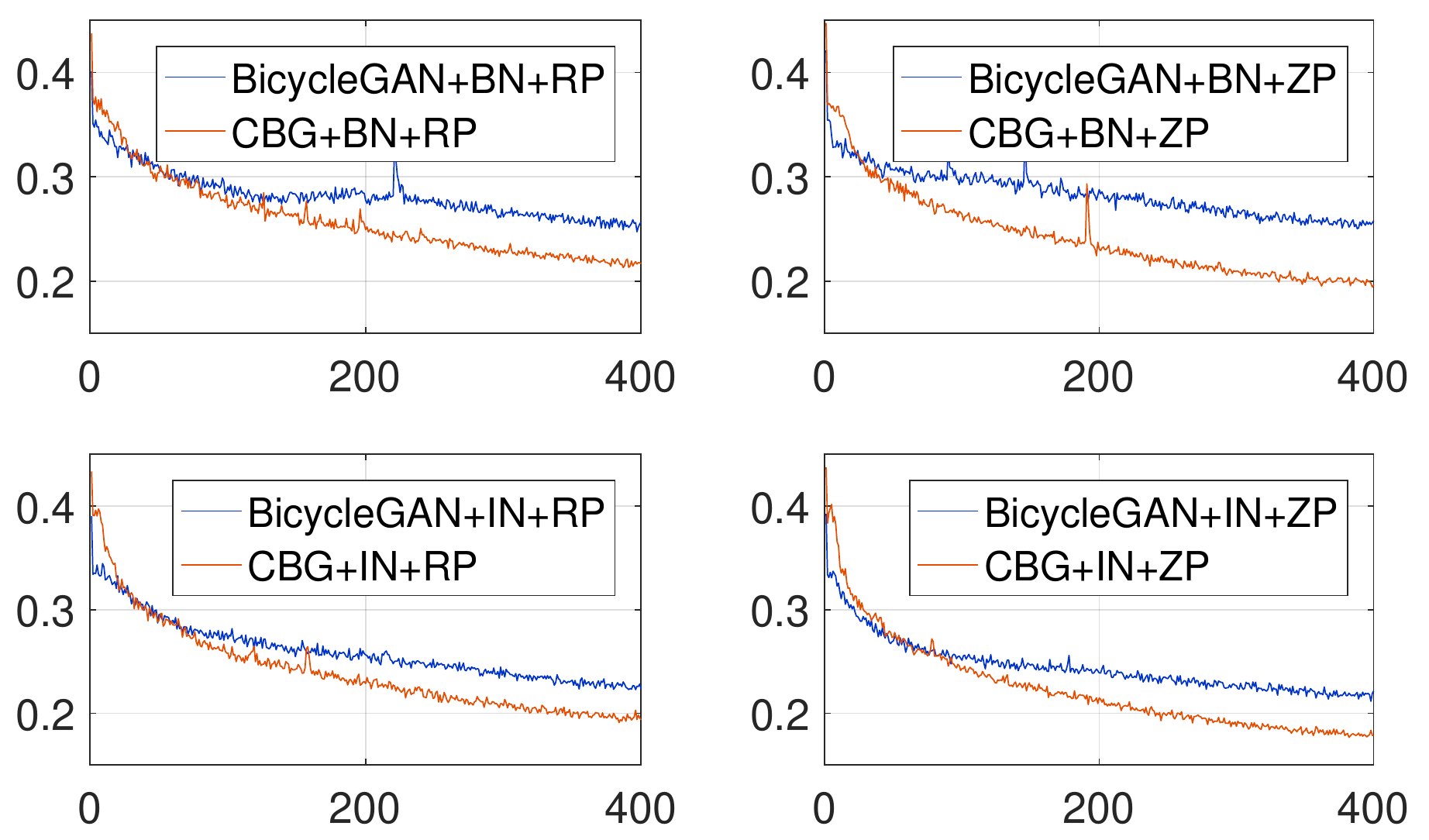}
	\caption{
		Comparison of convergence rate on \inlinecode{label2photo}.
%		The reconstruction loss of \inlinecode{label2photo} in training stage.
		The horizontal axis shows the iteration and the vertical axis represents the reconstruction loss.
	}
	\label{fig:convergence}
\end{figure}

In this section,
we give empirical evidence to demonstrate that the proposed CBN can accelerate the training of the multi-mapping model.
By comparing the reconstruction loss of BicycleGAN with and without CBN in Fig.~\ref{fig:convergence},
we can observe that CBN accelerates the model convergence and reduces the final reconstruction loss.
The reason is that CBN can reduce the internal covariate shift of the generator as mentioned in Section~\ref{sec:cbn}.

\section{Conclusions and Future Work} % (fold)
\label{sec:conclusions}
In this paper,
we study the latent code injection in the convolution neural network
and show the potential problems through different multi-mapping models,
\eg StarGAN, BicycleGAN, and pix2pix.
By decomposing the convolution output,
we show that the latent code controls the target mapping by providing offsets to the output feature maps.
%we show that the latent code provides the constant offsets to affect the mean value of output feature maps.
With further analysis,
we find the \textit{mapping inconsistency} and \textit{mapping indistinguishability} in the existing methods.
To overcome these problems,
we propose the \textit{consistency within diversity criteria} as a guide to designing the multi-mapping model.
Based on the criteria,
we propose central biasing normalization to replace the existing latent code injection strategy.
There are many advantages of our method: 
\begin{itemize}
	\item{It solves the mapping inconsistency and indistinguishability caused by normalization.}
	\item{It is insensitive to the selection of network structure, \eg the length latent code, padding strategies, or normalization operations.}
	\item{It has fewer parameters and faster convergence than common LCI model.}
	\item{It is easy to integrate into the latent-code-based models.}
\end{itemize}

Among a variety of multi-mapping translation tasks,
CBN provides a solid improvement over baselines.
Our CBN can be used in a variety of image processing tasks that need to inject auxiliary information into the convolution neural network,
such as arbitrary image stylization, multi-view generation, and diverse image inpainting.
For more specific applications,
we refer the reader to our related works SingleGAN\footnote{\href{https://github.com/Xiaoming-Yu/SingleGAN}{https://github.com/Xiaoming-Yu/SingleGAN}}~\cite{yu2018singlegan} and DMIT\footnote{\href{https://github.com/Xiaoming-Yu/DMIT}{https://github.com/Xiaoming-Yu/DMIT}}~\cite{yu2019multi}.
In our future work,
we will investigate whether CBN can help extend the representation ability of the discriminative model.
For example,
whether the domain-related information can be injected into the network for tackling the dataset bias problem~\cite{torralba2011unbiased} will be explored.
Besides,
the bias provided by central biasing normalization is indiscriminate for the entire feature map,
even though the transfer is non-global, \eg facial expression transfer.
Combining our method with spatial attention~\cite{mejjati2018unsupervised,yang2019show} or saliency mechanisms~\cite{fan2018salient,zhao2019EGNet,zhao2019contrast,wang2017video} would be an interesting work to improve the local transformation.
Finally,
we believe that this work is valuable for studying the multi-mapping translation.
Further exploration will allow the proposed method to be more universal and effective in different applications.

\ifCLASSOPTIONcaptionsoff
  \newpage
\fi

\bibliographystyle{IEEEtran}
%\bibliography{./main}
% Generated by IEEEtran.bst, version: 1.14 (2015/08/26)

\end{document}